\documentclass[11pt]{article}

\PassOptionsToPackage{table}{xcolor}
\usepackage[final]{acl}

\usepackage{times}
\usepackage{latexsym}

\usepackage[T1]{fontenc}

\usepackage[utf8]{inputenc}

\usepackage{microtype}

\usepackage{inconsolata}

\usepackage{graphicx}

\usepackage{booktabs}
\usepackage{makecell}
\usepackage{pifont}
\usepackage{multirow}
\usepackage{graphicx}
\usepackage{amsmath,amssymb}
\usepackage{array}
\usepackage{tabularx}
\usepackage{algorithm}
\usepackage{algpseudocode}
\usepackage{dsfont}
\usepackage[table]{xcolor}
\usepackage{enumitem}

\definecolor{oursbg}{gray}{0.95}

\newcommand{\score}[2]{#1{\scriptsize$\pm$#2}}
\usepackage{listings}

\lstdefinestyle{promptstyle}{
    basicstyle=\small\ttfamily,
    columns=fullflexible,
    breaklines=true,
    breakatwhitespace=false,
    keepspaces=true,
    showstringspaces=false,
    numbers=none,
    frame=none,
    linewidth=\linewidth,
    xleftmargin=0pt,
    xrightmargin=0pt,
    aboveskip=4pt,
    belowskip=4pt
}

\newcommand{\cmark}{\textcolor{green!60!black}{\ding{51}}}
\newcommand{\xmark}{\textcolor{red!70!black}{\ding{55}}}

\title{\textsc{DoC}: Learning to \underline{D}efer \underline{o}r \underline{C}ommit for Timely, Accurate, and Stable Conversational Diagnosis Prediction}

\author{
  Xueren Ge\textsuperscript{1}, Anthony Cortez\textsuperscript{2}, Homa Alemzadeh\textsuperscript{1} \\
  \textsuperscript{1}School of Engineering and Applied Sciences, University of Virginia, Charlottesville, VA, USA \\
  \textsuperscript{2}School of Medicine, University of Virginia, Charlottesville, VA, USA \\
  \texttt{\{zar8jw, aec3gp, ha4d\}@virginia.edu}
}

\begin{document}
\maketitle
\begin{abstract}
Conversational diagnosis prediction requires accurate, early, and stable predictions from ongoing clinical dialogue, yet dialogue-level labels do not specify when to commit. We propose a two-phase framework for learning deferral and diagnosis decisions. First, iterative commitment-boundary estimation constructs \textsc{Defer}/diagnosis supervision without annotated commitment times. Second, \textsc{Defer}-aware supervised fine-tuning initializes a policy, refined through reinforcement learning with trajectory-level rewards for timeliness, first-commit accuracy, and post-commit stability. On EMSDialog and MediTOD, \textsc{DoC} commits early, achieving the highest first-commit accuracy among post-trained methods (76.7\%/51.2\%) and the lowest post-commit error overall (16.4\%/48.3\%).

\end{abstract}

\section{Introduction}

Conversational diagnosis prediction (CDP) aims to identify a patient's likely condition during an ongoing medical conversation~\cite{ge-etal-2026-emsdialog}. Timely diagnostic suggestions can help clinicians prioritize assessment and plan next steps, particularly in time-critical settings such as emergency medical services (EMS)~\cite{10.1109/IoTDI61053.2024.00012}. In these encounters, evidence emerges incrementally across exchanges among the patient, clinicians, bystanders, and other responders. A diagnostic assistant~\cite{shu2019behavior} must therefore reason over incomplete information in real time to decide when the available evidence justifies committing to a prediction versus when to defer and wait for more evidence, and to keep its predictions stable as the conversation continues.

Training such an assistant poses two main challenges. 
First, the evidence supporting a diagnosis accumulates as the conversation unfolds, but the dialogue-level label does not indicate when that diagnosis first became reliably inferable. Using the final diagnosis as the target at every dialogue prefix (the conversation observed so far) can encourage premature and often incorrect commitments~\cite{kementchedjhieva-sogaard-2021-dynamic}.
Second, commitment timeliness, accuracy, and stability cannot be optimized independently: deferring longer delays assistance, committing earlier risks an incorrect first prediction, and an incorrect prediction forces a choice between correcting it and remaining consistent~\cite{madureira-schlangen-2020-incremental}. Training must therefore teach the model (i) \textit{when to commit}, (ii) \textit{what to predict}, and (iii) \textit{how to sustain correctness }afterward, accounting for the consequences of its decisions across the dialogue.

However, no existing approach jointly learns all three: (i) \emph{Selective prediction} learns to abstain via self-evaluation~\cite{chen-etal-2023-adaptation}, refusal-aware supervision~\cite{zhang-etal-2024-r}, calibrated uncertainty~\cite{tomani2024uncertainty}, or prompting~\cite{machcha-etal-2026-knowing}, but on static inputs, where abstention is a final refusal with no notion of \emph{when} evidence becomes sufficient. (ii) \emph{Early prediction} optimizes the accuracy--delay trade-off on a growing input~\cite{liu-etal-2022-improving,wang-etal-2024-esdm,Yuan_Singh_2023,zhang-etal-2025-forecasting}, but typically targets binary outcomes and treats the emitted decision as terminal, leaving later predictions unconstrained. (iii) \emph{Within CDP}, EMSDialog~\cite{ge-etal-2026-emsdialog} evaluates accuracy, earliness, and stability, but derives deferral from a fixed confidence threshold rather than learning it.  CaRT~\cite{liu2025cartteachingllmagents} treats stopping as terminal and delegates diagnosis to an external model, rather than learning both commitment and diagnosis. Question-asking agents~\cite{li2024mediq,lai2026doctorr} actively elicit information before diagnosing, assuming control over the conversation that a passive assistant in a multi-party EMS encounter
lacks. These gaps motivate learning the full sequence of deferral and diagnosis decisions, optimized for the correctness of predictions after commitment (Table~\ref{tab:related_work_comparison}).

\begin{table*}[t!]
\centering
\small
\setlength{\tabcolsep}{1.8pt}
\renewcommand{\arraystretch}{1}

\begin{tabular}{@{}p{2.05cm}lccclccc@{}}
\toprule
\textbf{Task} & \textbf{Work} & \textbf{Input} & \textbf{Output}
& \textbf{Defer} & \textbf{Defer via} & \textbf{Sequence}
& \textbf{Stability} & \textbf{Assistive} \\
\midrule

\multirow{4}{2.05cm}{\shortstack[l]{\textbf{Selective}\\\textbf{Prediction}}}
& ASPIRE~\cite{chen-etal-2023-adaptation}
& \multirow{4}{*}{Static}
& \multirow{4}{*}{\hspace*{1mm}Multi-class\hspace*{-1mm}}
& \cmark & SFT & \xmark & \xmark & -- \\

& R-Tuning~\cite{zhang-etal-2024-r}
& & &
\cmark & SFT & \xmark & \xmark & -- \\

& \cite{tomani2024uncertainty}
& & &
\cmark & Uncertainty & \xmark & \xmark & -- \\

& MedAbstain~\cite{machcha-etal-2026-knowing}
& & &
\cmark & Prompting & \xmark & \xmark & -- \\

\midrule

\multirow{3}{2.05cm}{\shortstack[l]{\textbf{Early}\\\textbf{Prediction}}}
& \cite{Yuan_Singh_2023}
& \multirow{3}{*}{Dynamic}
& \multirow{3}{*}{Binary}
& \xmark & -- & \cmark & \xmark & -- \\

& ESDM~\cite{wang-etal-2024-esdm}
& & &
\cmark & SFT & \cmark & \xmark & -- \\

& \cite{zhang-etal-2025-forecasting}
& & &
\xmark & -- & \cmark & \xmark & -- \\

\midrule

\multirow{5}{2.05cm}{\shortstack[l]{\textbf{Conversational}\\\textbf{Diagnosis}\\\textbf{Prediction}}}
& MediQ~\cite{li2024mediq}
& \multirow{5}{*}{Dynamic}
& \multirow{5}{*}{\hspace*{1mm}Multi-class\hspace*{-1mm}}
& \cmark & Prompting & -- & \xmark & \xmark \\

& CaRT~\cite{liu2025cartteachingllmagents}
& & &
\cmark & SFT & \cmark & \xmark & \cmark \\

& EMSDialog~\cite{ge-etal-2026-emsdialog}
& & &
\cmark & Uncertainty & \cmark & \cmark & \cmark \\

& Doctor-R1~\cite{lai2026doctorr}
& & &
\cmark & RL & -- & \xmark & \xmark \\

& \textbf{DoC (Ours)}
& & &
\cmark & \textbf{SFT + RL} & \cmark & \cmark & \cmark \\

\bottomrule
\end{tabular}
\vspace{-0.5em}
\caption{
Related work in selective, early, and conversational diagnosis prediction.
\textbf{Sequence}: decision sequence throughout the dialogue;
\textbf{Stability}: post-commit behavior;
\textbf{Assistive}: clinician-led setting.
}
\label{tab:related_work_comparison}
\vspace{-1em}
\end{table*}

\begin{figure*}[t]
    \centering
    \includegraphics[width=\textwidth]
    {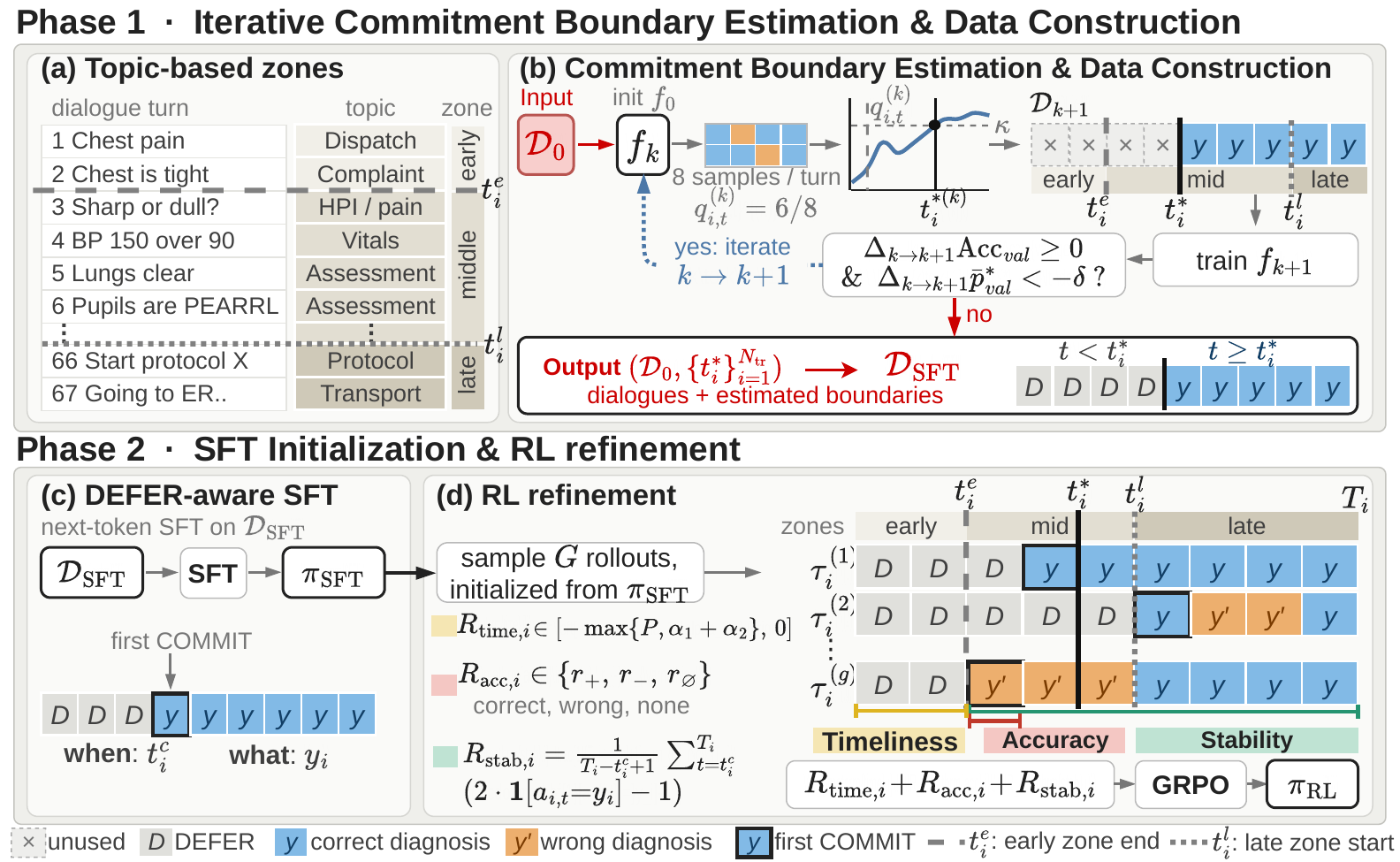}
    \caption{Overview of our framework. \textbf{Phase 1} constructs supervision via (a) topic-based early-evidence boundaries and (b) iterative commitment-boundary estimation. \textbf{Phase 2} trains the policy via (c) \textsc{Defer}-aware SFT and (d) RL with timeliness, first-commit accuracy, and post-commit stability rewards.
    }
    \label{fig:pipeline}
    \vspace{-1em}
\end{figure*}

To address these gaps, we propose \textsc{DoC}, a two-phase training framework for conversational diagnosis prediction (Figure~\ref{fig:pipeline}). The first phase iteratively estimates commitment boundaries and constructs explicit supervision for deferral and diagnosis. The second phase uses this supervision to initialize a diagnostic policy by supervised fine-tuning (SFT), then refines its decisions over complete dialogue trajectories by reinforcement learning (RL) to encourage timely, correct commitments and stability afterward.
Our contributions are threefold:
\begin{itemize}[itemsep=3pt, topsep=3pt, parsep=0pt, partopsep=0pt]
    \item We develop an iterative procedure that estimates dialogue-specific commitment boundaries from topic-based early-evidence boundaries and a diagnosis model's prefix-level correctness, providing explicit \textsc{Defer}/diagnosis supervision without manual annotation.
    \item We develop a training strategy that initializes the policy with \textsc{Defer}-aware SFT and refines it by RL with trajectory-level rewards for commitment timeliness, first-commit accuracy, and post-commit stability, allowing the policy to improve beyond initial supervision.
    \item We demonstrate on EMSDialog~\cite{ge-etal-2026-emsdialog} and MediTOD~\cite{saley-etal-2024-meditod} that \textsc{DoC} achieves a strong accuracy--earliness trade-off among post-trained methods (first-correct earliness 66.1/ 38.1; first-commit accuracy 76.7\%/51.2\%) and the lowest post-commit error of all methods (16.4\%/48.3\%).
\end{itemize}

\section{Related Work}
\paragraph{Selective prediction.} Selective prediction allows a model to abstain when its prediction may be unreliable.~ASPIRE~\cite{chen-etal-2023-adaptation} adapts an LLM for task prediction and trains a separate self-evaluation component to estimate whether its answer is correct.~R-Tuning~\cite{zhang-etal-2024-r} constructs refusal-aware supervision and fine-tunes models to distinguish questions they can answer from those requiring an uncertain response. \citet{tomani2024uncertainty} use uncertainty estimates to decide whether to return an answer, while MedAbstain~\cite{machcha-etal-2026-knowing} studies prompting-based abstention under clinical uncertainty.~As shown in Table~\ref{tab:related_work_comparison}, these methods  make a single answer-or-abstain decision on a static input.~In CDP, a model makes a sequence of defer-or-diagnose decisions over successive dialogue prefixes, waits for more evidence when needed, and continues predicting after committing.
\paragraph{Early Prediction.} Early prediction seeks to infer an outcome from partially observed sequential data. On clinical notes, LuPIET~\cite{liu-etal-2022-improving} distills a teacher using later text as privileged information into a student restricted to the early window. In the dialogue domain, ESDM~\cite{wang-etal-2024-esdm} detects depression early by jointly learning what to predict from partial observations and when to classify, balancing accuracy and delay. Other works forecast conversation derailment from evolving prefixes~\cite{chang-danescu-niculescu-mizil-2019-trouble,Yuan_Singh_2023,zhang-etal-2025-forecasting}. Although these methods model dynamic inputs and prediction earliness, they predominantly consider binary outcomes and treat the emitted decision as terminal. CDP extends this setting to multiclass diagnosis, where the model must commit early and remain correct afterward (Table~\ref{tab:related_work_comparison}).

\paragraph{Conversational Diagnosis Prediction.} Recent works on interactive clinical reasoning study how models gather sufficient information before answering. MediQ~\cite{li2024mediq} uses an abstention module to decide whether an expert agent should answer or ask the patient a follow-up question. Doctor-R1~\cite{lai2026doctorr} applies RL to optimize strategic clinical inquiry and final diagnostic decisions. These systems let the model control the interaction by selecting questions and terminate with a final answer. Our setting is instead \textit{clinician-led}: the model \textit{passively} observes an ongoing medical conversation and continuously determines whether to \textsc{Defer} or \textsc{Commit}. 

EMSDialog~\cite{ge-etal-2026-emsdialog} and CaRT~\cite{liu2025cartteachingllmagents} are the most closely related works, evaluating diagnosis predictions across conversation turns. However, EMSDialog obtains deferral from a fixed confidence threshold at inference-time and CaRT learns only when to stop gathering evidence. Neither learns a policy that jointly optimizes accuracy, timeliness, and post-commit stability. \textsc{DoC} addresses this gap by learning when to commit, what to predict, and how to remain correct afterward over complete dialogue trajectories.

\section{Problem Formulation}
\label{sec:problem_formulation}

CDP requires an assistive model to decide \textit{when} and \textit{what} to predict as a clinician-led dialogue unfolds. For dialogue $i$ with ground-truth diagnosis $y_i\in\mathcal{Y}$, let $x_{i,1},\ldots,x_{i,T_i}$ be the successively accumulating prefixes given to the model, taken at a fixed utterance stride (Section~\ref{sec:setup}), where $x_{i,T_i}$ is the full dialogue. Based only on $x_{i,t}$, the model selects $a_{i,t}\in\mathcal{A}=\mathcal{Y}\cup\{\textsc{Defer}\}$. Here, \textsc{Defer} withholds a diagnosis pending further evidence, while selecting a diagnosis constitutes a \textsc{Commit}. The first commitment occurs at
\vspace{-0.2em}
\begin{equation}
t_i^{\mathrm{c}}
=
\min\{t:a_{i,t}\neq\textsc{Defer}\},
\label{eq:first_commit}
\vspace{-0.5em}
\end{equation}
with $t_i^{\mathrm{c}} = T_i + 1$ if the model never commits. The model continues predicting throughout the dialogue, producing the action trajectory $(a_{i,1},\ldots,a_{i,T_i})$, of which
the portion from the first commitment onward, $(a_{i,t_i^{\mathrm{c}}},\ldots,a_{i,T_i})$, is the post-commit trajectory.
Given $N_{\mathrm{tr}}$ fully observed training dialogues $\mathcal{D}_0=\{(x_{i,T_i},y_i)\}_{i=1}^{N_{\mathrm{tr}}}$, we seek a policy $\pi_\theta(a\mid x_{i,t})$ that jointly balances commitment timeliness, first-commit accuracy, and post-commit stability. Full notation is in Appendix Table~\ref{tab:notation}. %

\begin{table*}[t]
\centering
\small
\setlength{\tabcolsep}{3pt}
\renewcommand{\arraystretch}{1.15}
\setlength{\heavyrulewidth}{1pt}
\begin{tabular*}{\textwidth}{@{\extracolsep{\fill}}llcccc@{}}
\toprule
& & \multicolumn{3}{c}{\textbf{Commit at $t_i^{\mathrm{c}}\leq T_i$}}
& \textbf{Never commit} \\
\cmidrule(lr){3-5}\cmidrule(l){6-6}
\textbf{Reward} & \textbf{First commit}
& $t_i^{\mathrm{c}}\leq t_i^{\mathrm{e}}$
& $t_i^{\mathrm{e}}< t_i^{\mathrm{c}}\leq t_i^*$
& $t_i^*<t_i^{\mathrm{c}}\leq T_i$
& $t_i^{\mathrm{c}}>T_i$ \\
\midrule
\multirow{2}{*}[-1.6ex]{$\boldsymbol{R_{\mathrm{time},i}}$}
& Correct\rule[-3.3ex]{0pt}{7ex}
& \multirow{2}{*}[-1.6ex]{$-P$}
& $-\alpha_1\dfrac{t_i^{\mathrm{c}}-t_i^{\mathrm{e}}}{\max\{t_i^*-t_i^{\mathrm{e}},1\}}$
& $-\alpha_1-\alpha_2
   \dfrac{t_i^{\mathrm{c}}-t_i^*}{\max\{T_i-t_i^*,1\}}$
& \multirow{2}{*}[-1.6ex]{$-\alpha_1-\alpha_2\dfrac{T_i-t_i^*}{\max\{T_i-t_i^*,1\}}$} \\
\cmidrule[0.25pt](r){2-2}\cmidrule[0.25pt](lr){4-5}
& Incorrect
&
& $0$
& $0$
& \\
\midrule
\multirow{2}{*}{$\boldsymbol{R_{\mathrm{acc},i}}$}
& Correct
& \multicolumn{3}{c}{$r_{+}$}
& \multirow{2}{*}{$r_{\varnothing}$} \\[-0.3ex]
\cmidrule[0.25pt](r){2-2}\cmidrule[0.25pt](lr){3-5}
& Incorrect
& \multicolumn{3}{c}{$r_{-}$}
& \\
\midrule
$\boldsymbol{R_{\mathrm{stab},i}}$
& Either\rule[-2.6ex]{0pt}{6.2ex}
& \multicolumn{3}{c}{$\dfrac{\sum_{t=t_i^{\mathrm{c}}}^{T_i}
   \bigl(2\cdot \mathds{1} [a_{i,t}=y_i]-1\bigr)}{T_i-t_i^{\mathrm{c}}+1}$}
& $0$ \\
\bottomrule
\end{tabular*}
\vspace{-0.5em}
\caption{
Reward definitions (Timeliness, Accuracy, Stability) for a single rollout.
}
\label{tab:reward_cases}
\vspace{-1em}
\end{table*}

\section{Methodology}
\label{sec:method}
To address the challenges of missing commitment-time supervision and jointly optimizing timeliness, accuracy, and stability, we propose a two-phase training framework. As shown in Figure~\ref{fig:pipeline}, Phase~1 constructs \textsc{Defer}/diagnosis supervision through topic-based boundary identification and iterative commitment-boundary estimation. Phase~2 trains the policy on this supervision with SFT, then refines it with RL using trajectory-level rewards.

\subsection{Early-Evidence Boundary Identification}
Early in a dialogue, the available information fits many diagnoses, so a correct prediction may reflect class priors or surface cues rather than evidence. We therefore use utterance-level topic annotations to mark an information-sparse \textit{early zone}, within which commitment boundaries are not placed (Section~\ref{sec:commit_boundary_estimation}) and
commitments are penalized (Section~\ref{sec:timeliness}). %
Let $z_{i,t}$ be the topic of the last utterance in $x_{i,t}$ and $\mathcal{Z}_{\mathrm{early}}$ the set of early-zone topics (e.g., Dispatch, Introduction, Chief Complaint in EMS; Appendix~\ref{appendix:conv_zones}). The early-evidence boundary $t_i^{\mathrm{e}}$ is the last prefix ending in an early-zone topic:
\vspace{-0.5em}
\begin{equation}
\vspace{-0.5em}
t_i^{\mathrm{e}}
=
\max\left\{
t:z_{i,t}\in\mathcal{Z}_{\mathrm{early}}
\right\}.
\label{eq:early_boundary}
\end{equation}
We also mark $t_i^{\ell}$ as the last prefix before a clinician first states the ground-truth diagnosis (beginning of the \textit{late zone} in Figure~\ref{fig:pipeline}b and d), which is used only to evaluate lateness (Appendix~\ref{app:macro_f1_lateness_edit_overhead}).

\subsection{Commitment Boundary Estimation}
\label{sec:commit_boundary_estimation}

Dialogue-level labels do not specify when a diagnosis becomes reliably inferable. We define this point empirically, as the earliest prefix after $t_i^{\mathrm{e}}$ from which a diagnosis model (\emph{labeler} $f_k$) consistently predicts $y_i$. Inspired by self-training~\cite{He2020Revisiting}, we iteratively estimate the commitment boundary $t_i^*$ from a labeler's prefix-level correctness, retraining it on prefixes from each boundary onward (Figure~\ref{fig:pipeline}b, Algorithm~\ref{alg:boundary_estimation}).

\paragraph{Boundary from prefix correctness.} 
We first train a diagnosis model on the complete training dialogues and use it as the
initial labeler $f_0$. At each iteration $k$, we estimate boundaries separately on
the training and validation splits. Training boundaries define prefix supervision, while validation boundaries guide stopping. For each prefix $x_{i,t}$, we sample $M$ diagnoses from the current labeler $f_k$ and compute their correctness rate:
\begin{equation}
\vspace{-0.5em}
q_{i,t}^{(k)}
=
\frac{1}{M}
\sum_{m=1}^{M}
\mathbf{1}\!\left[
\hat{y}_{i,t}^{(k,m)}=y_i
\right],
\label{eq:prefix_correctness}
\end{equation}
where $\hat{y}_{i,t}^{(k,m)}$ is the $m$-th sample. The commitment boundary at iteration $k$ is the earliest prefix after $t_i^{\mathrm{e}}$ from which this rate stays at or above $\kappa$:
\begin{equation}
t_i^{*(k)}
=
\min\left\{
t>t_i^{\mathrm{e}}:
q_{i,t'}^{(k)}\geq\kappa\ \ \forall\, t'\in[t,T_i]
\right\}.
\label{eq:pseudo_boundary}
\end{equation}
This excludes early-zone prefixes and filters transient correctness spikes. If none qualifies, we set $t_i^{*(k)}=T_i$ so every dialogue ends in a diagnosis.

\paragraph{Iterative refinement.} 
Using the training boundaries, we construct a diagnosis-only prefix dataset:
\begin{equation}
\mathcal{D}_{k+1}
=
\bigcup_{i=1}^{N_{\mathrm{tr}}}
\left\{
(x_{i,t},y_i):
t_i^{*(k)}\leq t\leq T_i
\right\}.
\label{eq:diagnosis_prefix_data}
\end{equation}
Earlier prefixes are omitted rather than labeled \textsc{Defer}. 
We train $f_{k+1}$ on $\mathcal{D}_{k+1}$ and re-estimate boundaries, accepting an update only if validation accuracy does not decrease and the mean normalized boundary advances sufficiently:

\begin{equation}
\begin{aligned}
\mathrm{Acc}_{\mathrm{val}}(f_{k+1})
&\geq \mathrm{Acc}_{\mathrm{val}}(f_k),\\
\bar{p}_{\mathrm{val}}^{*(k+1)}-\bar{p}_{\mathrm{val}}^{*(k)}
&<-\delta,
\end{aligned}
\label{eq:pseudo_convergence}
\end{equation}
where
$\bar{p}_{\mathrm{val}}^{*(k)}
=\frac{1}{N_{\mathrm{val}}}\sum_{i=1}^{N_{\mathrm{val}}}t_i^{*(k)}/T_i$ is the mean normalized validation boundary, and $N_{\mathrm{val}}$ is the number of validation dialogues. If either condition fails, we stop and retain the training-set boundaries $\{t_i^*\}_{i=1}^{N_{\mathrm{tr}}}$
from the last accepted iteration.

\paragraph{Supervision construction.}
The final boundaries define a target action for each dialogue prefix:
\begin{equation}
\vspace{-0.2em}
a_{i,t}^*
=
\begin{cases}
\textsc{Defer}, & t<t_i^*,\\
y_i, & t\geq t_i^*.
\end{cases}
\label{eq:sft_target}
\end{equation}
This yields a dataset
$\mathcal{D}_{\mathrm{SFT}}
=\bigcup_{i=1}^{N_{\mathrm{tr}}}
\{(x_{i,t},a_{i,t}^*):t=1,\ldots,T_i\}$
for SFT initialization.

\subsection{\textsc{Defer}-aware Supervised Fine-tuning}
\label{sec:sft}

We initialize the SFT policy $\pi_{\mathrm{SFT}}$ by minimizing the negative log-likelihood of targets in $\mathcal{D}_{\mathrm{SFT}}$:
\vspace{-0.5em}
\begin{equation}
\mathcal{L}_{\mathrm{SFT}}
=
-\frac{1}{\sum_i T_i}
\sum_{i=1}^{N_{\mathrm{tr}}}
\sum_{t=1}^{T_i}
\log\pi_\theta
\!\left(
a_{i,t}^*\mid x_{i,t}
\right).
\label{eq:sft_loss}
\end{equation}

\subsection{Reinforcement Learning Refinement}
\label{sec:rl}

\textsc{Defer}-aware SFT imitates the hard target $a_{i,t}^*$ at each prefix independently. To move beyond fixed prefix-level targets, we refine $\pi_{\mathrm{SFT}}$ using rewards over complete prediction trajectories. Starting from $\pi_{\mathrm{SFT}}$, we sample $G$ trajectories per dialogue from the current policy:
\begin{equation}
\tau_i^{(g)}
=
\left(
a_{i,1}^{(g)},\ldots,a_{i,T_i}^{(g)}
\right),
\qquad
g=1,\ldots,G.
\end{equation}
Each trajectory is evaluated using the three rewards defined below and summarized in Table~\ref{tab:reward_cases}. 

\subsubsection{Timeliness Reward}
\label{sec:timeliness}

The timeliness reward encourages earlier correct commitments (Table~\ref{tab:reward_cases}, row $R_{\mathrm{time},i}$; Appendix Figure~\ref{fig:reward_appendix}a). Commitments at or before $t_i^{\mathrm{e}}$ receive $-P$ regardless of correctness, discouraging premature guesses. For later correct commitments, the delay cost grows linearly by $\alpha_1$ %
up to $t_i^*$ and by a further $\alpha_2$ up to $T_i$. Incorrect commitments after $t_i^{\mathrm{e}}$ receive $0$, giving an earlier wrong guess no advantage over a later one. Non-commitment incurs the maximum delay penalty to discourage indefinite deferral.

\subsubsection{First-Commit Accuracy Reward}
\label{sec:first_commit_reward}

We score only the first committed diagnosis: $r_{+}$ if it is correct, $r_{-}$ if it is incorrect, and $r_{\varnothing}$ if the policy never commits (Table~\ref{tab:reward_cases}, row $R_{\mathrm{acc},i}$; Appendix Figure~\ref{fig:reward_appendix}b). %
We choose $r_{-}<r_{\varnothing}<0<r_{+}$, penalizing an incorrect
commitment more heavily than non-commitment, which prioritizes
diagnostic correctness while discouraging indefinite deferral.

\subsubsection{Post-Commit Stability Reward}
\label{sec:stability_reward}

For each prefix from the first commitment onward ($t_i^{\mathrm{c}}\leq t\leq T_i$), we assign $+1$ for a correct diagnosis and $-1$ for an incorrect diagnosis or \textsc{Defer}. The stability reward $R_{\mathrm{stab},i}$ is the average score over these prefixes, or $0$ if the policy never commits (Table~\ref{tab:reward_cases}, row $R_{\mathrm{stab},i}$; 
Appendix Figure~\ref{fig:reward_appendix}c). This reward encourages sustained correctness after commitment and discourages returns to \textsc{Defer}.

\subsubsection{Overall Objective}
\label{sec:rl_objective}

The total trajectory reward is
\begin{equation}
R_i = R_{\mathrm{time},i} + R_{\mathrm{acc},i} + R_{\mathrm{stab},i}.
\label{eq:total_reward}
\end{equation}
We use an on-policy GRPO variant~\cite{shao2024deepseekmathpushinglimitsmathematical} with one optimizer step per rollout batch, so the importance ratio is one and clipping is inactive when computing gradients. We compute
\(
A_i^{(g)}
=
(R_i^{(g)}-\bar{R}_i)/(\sigma_i+\eta)
\),
where $\bar{R}_i$ and $\sigma_i$ are the mean and standard deviation
of dialogue $i$'s $G$ rewards, and $\eta$ ensures numerical stability. We minimize
\begin{equation}
\begin{aligned}
&\mathcal{L}_{\mathrm{RL}}(\theta)
=-\mathbb{E}_{i,g}\!\left[
 A_i^{(g)}
 \sum_{t=1}^{T_i}
 \log\pi_\theta\!\left(
 a_{i,t}^{(g)}\mid x_{i,t}
 \right)
 \right]\\
&\quad+\beta\,\mathbb{E}_{i,g,t}\!\left[
 D_{\mathrm{KL}}\!\left(
 \pi_\theta(\cdot\mid x_{i,t})
 \,\|\,
 \pi_{\mathrm{SFT}}(\cdot\mid x_{i,t})
 \right)
 \right],
\end{aligned}
\label{eq:policy_loss}
\end{equation}
where $\beta$ controls the KL penalty toward the fixed SFT policy $\pi_{\mathrm{SFT}}$.

\section{Experimental Setup}
\subsection{Datasets}
We evaluate \textsc{DoC} on two medical dialogue datasets. \textbf{EMSDialog}~\citep{ge-etal-2026-emsdialog} comprises 4{,}411 synthetic EMS dialogues for training and 60 transcribed EMS simulations for testing; we merge adult and pediatric variants of its 43 protocol codes into 35 diagnosis labels. %
\textbf{MediTOD}~\citep{saley-etal-2024-meditod} contains doctor--patient conversations from medical history-taking examinations, released as raw transcripts with free-text diagnoses and no data split; %
we canonicalize the diagnoses to 33 labels and randomly split the 231 dialogues into 162/69 for training/testing. In all experiments, diagnosis names mentioned in the dialogues are masked to prevent label leakage (Appendix~\ref{app:data_prep}).

\subsection{Baselines}
We compare \textsc{DoC} with SOTA in-context baseline LLMs and the post-trained methods from Table~\ref{tab:related_work_comparison}. 

\noindent\textbf{In-context baselines.}
Following the MedAbstain~\cite{machcha-etal-2026-knowing}, we prompt Qwen3-1.7B/4B, Doctor-R1-8B~\cite{lai2026doctorr}, Claude Opus 5~\cite{anthropic2026opus5}, and GPT-6 Sol~\cite{openai2026gpt6sol} with thinking disabled (Appendix~\ref{app:prompt}).

\noindent\textbf{Post-trained methods.} All post-trained methods use Qwen3-1.7B~\cite{yang2025qwen3technicalreport}:

\noindent\textbf{ASPIRE}~\cite{chen-etal-2023-adaptation}
trains two LoRA adapters for diagnosis and correctness estimation; the latter decides whether to \textsc{Commit} or \textsc{Defer}.

\noindent\textbf{R-Tuning}~\cite{zhang-etal-2024-r}
uses pretrained-model correctness to construct answer and abstention examples for instruction tuning, mapping answers to \textsc{Commit} and refusals to \textsc{Defer}.

\noindent\textbf{ESDM}~\cite{wang-etal-2024-esdm}
jointly trains a partial-information classifier and a stopping module with classification, decision, and delay losses.

\noindent\textbf{CaRT}~\cite{liu2025cartteachingllmagents}
trains a termination model on counterfactual sufficient/insufficient information pairs while keeping the diagnosis model fixed.

\noindent\textbf{Static/Dynamic Train}~\cite{ge-etal-2026-emsdialog}
trains on complete dialogues or the last $5$ prefixes, respectively, and commits when the classifier's confidence exceeds 0.5 (Appendix~\ref{app:implementation}).

\begin{table*}[t]
\centering
\small
\setlength{\tabcolsep}{3pt}
\renewcommand{\arraystretch}{1.1}

\resizebox{\textwidth}{!}{%
\begin{tabular}{@{}l*{10}{c}@{}}
\toprule
\multirow{2}{*}{\textbf{Method}}
& \multicolumn{5}{c}{\textbf{EMSDialog}}
& \multicolumn{5}{c}{\textbf{MediTOD}} \\
\cmidrule(lr){2-6}
\cmidrule(lr){7-11}
& \textbf{Acc@Cmt}$\uparrow$
& \textbf{Acc@1.0}$\uparrow$
& \textbf{Earl.}$\uparrow$
& \textbf{Acc@Def}$\uparrow$
& \textbf{PC-Err}$\downarrow$
& \textbf{Acc@Cmt}$\uparrow$
& \textbf{Acc@1.0}$\uparrow$
& \textbf{Earl.}$\uparrow$
& \textbf{Acc@Def}$\uparrow$
& \textbf{PC-Err}$\downarrow$ \\
\midrule

\multicolumn{11}{@{}l}{\textbf{In-context}} \\

Qwen3-1.7B
& \score{10.6}{2.5}
& \score{4.4}{2.5}
& \score{5.4}{2.5}
& \score{50.3}{3.3}
& \score{86.6}{1.7}
& \score{15.9}{1.4}
& \score{14.0}{2.2}
& \score{5.5}{2.5}
& \score{43.8}{1.2}
& \score{88.0}{4.3} \\

Qwen3-4B
& \score{72.8}{1.9}
& \score{66.1}{1.0}
& \score{62.8}{4.0}
& \score{65.0}{2.3}
& \score{26.1}{2.9}
& \score{21.7}{1.4}
& \score{39.6}{1.7}
& \score{30.2}{2.7}
& \score{39.6}{3.3}
& \score{72.8}{4.4} \\

Doctor-R1-8B
& \score{64.4}{2.5}
& \score{59.4}{1.0}
& \score{69.9}{2.3}
& \score{69.7}{2.7}
& \score{28.2}{2.8}
& \score{32.4}{2.2}
& \score{41.5}{2.2}
& \score{28.5}{4.4}
& \score{48.0}{3.9}
& \score{72.3}{4.2} \\

Claude Opus 5
& \score{\textbf{82.8}}{1.0}
& \score{80.0}{1.7}
& \score{71.5}{1.9}
& \score{77.7}{1.6}
& \score{17.9}{1.2}
& \score{\textbf{53.6}}{1.4}
& \score{55.6}{0.8}
& \score{36.3}{1.2}
& \score{56.4}{1.5}
& \score{52.7}{0.5} \\

GPT-6 Sol
& \score{75.0}{1.7}
& \score{\textbf{81.1}}{1.0}
& \score{\textbf{75.5}}{1.6}
& \score{75.8}{1.7}
& \score{18.1}{1.1}
& \score{48.8}{0.8}
& \score{51.7}{0.8}
& \score{38.2}{1.6}
& \score{57.4}{1.2}
& \score{53.9}{1.3} \\

\midrule
\multicolumn{11}{@{}l}{%
  \textbf{Post-trained (Qwen3-1.7B)}} \\

ASPIRE
& \score{71.1}{2.5}
& \score{63.9}{2.5}
& \score{48.3}{3.8}
& \score{57.8}{5.2}
& \score{28.5}{4.8}
& \score{40.1}{2.2}
& \score{49.3}{1.4}
& \score{52.7}{5.5}
& \score{41.8}{4.3}
& \score{59.2}{5.0} \\

ESDM
& \score{67.2}{3.5}
& \score{70.0}{3.3}
& \score{52.8}{4.4}
& \score{72.5}{3.6}
& \score{30.7}{5.7}
& \score{35.3}{2.2}
& \score{36.2}{1.4}
& \score{31.9}{2.2}
& \score{37.0}{1.9}
& \score{65.2}{2.7} \\

R-Tuning
& \score{70.0}{3.3}
& \score{77.2}{2.5}
& \score{65.4}{2.8}
& \score{76.8}{3.1}
& \score{25.1}{3.1}
& \score{47.8}{1.4}
& \score{43.5}{2.5}
& \score{38.0}{2.1}
& \score{58.1}{2.8}
& \score{64.8}{2.9} \\

CaRT
& \score{61.1}{2.5}
& \score{74.4}{1.0}
& \score{37.0}{2.2}
& \score{56.3}{1.9}
& \score{27.1}{2.2}
& \score{35.7}{0.8}
& \score{57.0}{0.8}
& \score{38.5}{1.2}
& \score{40.4}{2.8}
& \score{60.6}{1.6} \\

Static Train
& \score{51.1}{3.5}
& \score{75.0}{1.7}
& \score{70.0}{3.4}
& \score{73.8}{3.6}
& \score{26.9}{4.0}
& \score{30.0}{3.0}
& \score{57.5}{0.8}
& \score{52.7}{1.4}
& \score{38.0}{2.1}
& \score{67.3}{3.1} \\

Dynamic Train
& \score{53.9}{1.0}
& \score{75.6}{1.0}
& \score{\underline{70.1}}{1.5}
& \score{74.9}{2.5}
& \score{24.7}{2.8}
& \score{40.6}{1.4}
& \score{58.0}{1.4}
& \score{\textbf{\underline{54.6}}}{3.3}
& \score{40.9}{3.8}
& \score{54.4}{1.2} \\

\addlinespace[3pt]
\rowcolor{oursbg}
\textbf{SFT (DoC)}
& \score{74.4}{1.0}
& \score{76.7}{1.7}
& \score{62.2}{0.9}
& \score{73.7}{1.8}
& \score{22.5}{2.8}
& \score{50.2}{0.8}
& \score{58.5}{1.7}
& \score{33.2}{2.4}
& \score{56.0}{2.5}
& \score{51.2}{3.5} \\

\rowcolor{oursbg}
\textbf{SFT + RL (DoC)}
& \score{\underline{76.7}}{1.7}
& \score{\underline{80.0}}{1.7}
& \score{66.1}{2.5}
& \score{\textbf{\underline{79.3}}}{2.6}
& \score{\textbf{\underline{16.4}}}{2.8}
& \score{\underline{51.2}}{0.8}
& \score{\textbf{\underline{59.4}}}{1.4}
& \score{38.1}{1.8}
& \score{\textbf{\underline{61.8}}}{2.2}
& \score{\textbf{\underline{48.3}}}{1.7} \\

\bottomrule
\end{tabular}%
}
\vspace{-0.5em}
\caption{
Main results on EMSDialog and MediTOD.
\textbf{Bold} indicates the best mean score overall;
\underline{underlining} indicates the best among the evaluated
open-weight models. Closed-source models are included for reference.
}
\label{tab:main_results}
\vspace{-0.5em}
\end{table*}

\subsection{Evaluation Metrics}
\label{sec:evaluation_metrics}
\textbf{Acc@Commit (Acc@Cmt)} ($\uparrow$) measures the accuracy of the first committed diagnosis, with non-committing diagnoses counted as incorrect. \\
\textbf{Acc@$\rho$} ($\uparrow$) measures diagnosis accuracy after observing a fixed fraction $\rho$ of each dialogue, enabling comparison at matched relative progress and reducing length-related bias~\cite{liu2025cartteachingllmagents}. Acc@1.0 evaluates on the complete dialogue. \\
\textbf{Earliness (Earl.)} ($\uparrow$) measures first-correct earliness \citep{gupta2020approaches}, defined as $1-t_i^{\mathrm{fc}}/T_i$, where $t_i^{\mathrm{fc}}$ indexes the first prefix with a correct prediction; it is zero if no correct prediction occurs. \\
\textbf{Deferral accuracy (Acc@Def)} ($\uparrow$) evaluates deferral decisions using Abstain Accuracy~\cite{feng2024don}. For each prefix, we obtain a forced diagnosis by re-decoding the same prompt with \textsc{Defer} disabled. Acc@Def measures the proportion of decisions that defer when the forced diagnosis is incorrect or commit when it is correct.\\
\textbf{Post-commit error rate (PC-Err)} ($\downarrow$) is the post-commit analogue of streaming prediction error \citep{kaushal2023efficient}: the proportion of decisions from the first commitment onward (inclusive) that fail to output the ground-truth diagnosis, computed over dialogues with a commitment. \\
Further details are provided in Appendix~\ref{app:evaluation_metrics}.

\subsection{Implementation Details}
\label{sec:setup}

We fully fine-tune Qwen3-1.7B for both the labelers and the policy, with AdamW~\cite{loshchilov2018decoupled}. We reserve 20\% of the training dialogues for validation. In training and evaluation, we take a prefix every two utterances (stride $=2$), always including the
complete dialogue as the final prefix. Boundary estimation uses $M=8$ samples per prefix, $\kappa=0.75$, and $\delta=0.005$. Reward parameters are $P=1$, $\alpha_1=1$, $\alpha_2=1$, and $(r_+,r_-,r_{\varnothing})=(1,-4.5,-1.5)$. RL uses $G=8$ trajectories per dialogue and a KL coefficient $\beta=0.005$. Table~\ref{tab:main_results} reports the mean $\pm$ sample standard deviation over three runs (seeds 1, 2, 42). All experiments run on NVIDIA A100 GPUs; full settings are in Appendix~\ref{app:implementation}.

\section{Results and Analysis}

We investigate the following research questions:
\textbf{RQ1}: How does \textsc{DoC} compare with
existing methods in accuracy, earliness,
and stability?\\
\textbf{RQ2}: How do boundary estimation, SFT initialization, and each RL reward contribute?\\
\textbf{RQ3}: How does \textsc{DoC} compare with a clinician, and does the clinician find its predictions helpful?
\subsection{RQ1: Comparison with Baselines}
\label{sec:rq1}

\begin{figure}[t]
    \centering
    \includegraphics[width=\columnwidth]
    {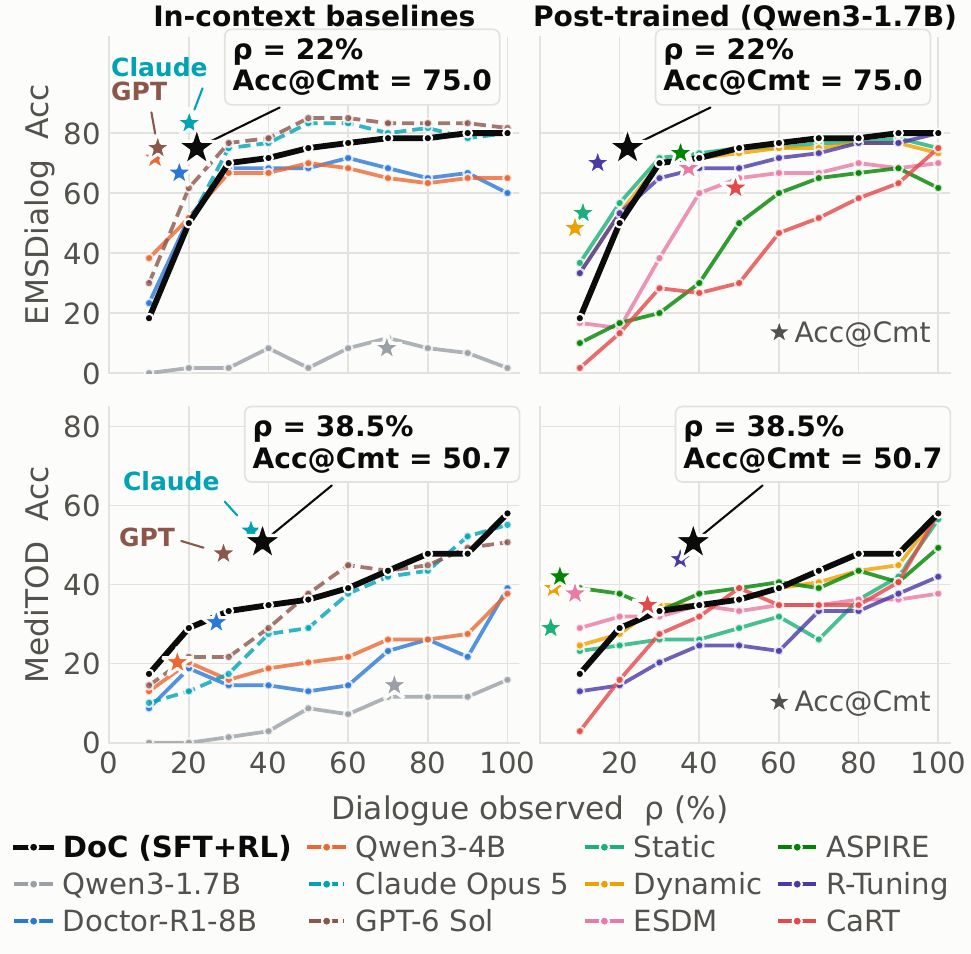}
    \vspace{-1.5em}
    \caption{
    Model's diagnosis accuracy at the fixed dialogue progress
    (Acc@$\rho$, seed 42).
    }
    \label{fig:acc_vs_commit}
    \vspace{-1.2em}
\end{figure}

\noindent\textbf{DoC achieves a strong balance among accuracy, earliness, and post-commit stability.} Table~\ref{tab:main_results} shows that \textsc{DoC} achieves the best Acc@Commit among post-trained methods on EMSDialog and MediTOD (76.7\% and 51.2\%), alongside the lowest PC-Err among all evaluated methods (16.4\% and 48.3\%). Figure~\ref{fig:acc_vs_commit} further shows that these accurate first commitments occur at only 22\% and 38.5\% of dialogue progress. At matched dialogue progress, \textsc{DoC} remains competitive on EMSDialog and achieves higher accuracy than most baselines on MediTOD. It also attains the highest Macro-F1@Commit among
post-trained methods and the lowest edit overheads overall, with no late commitments in 92.8\%/96.6\% of EMSDialog/MediTOD dialogues (Appendix~\ref{app:macro_f1_lateness_edit_overhead}).
These results show that \textsc{DoC} commits accurately from partial observations and maintains low post-commit error.

\noindent\textbf{DoC remains competitive with strong in-context models.} Despite using a 1.7B backbone, \textsc{DoC} achieves competitive performance in Acc@Commit compared to Claude Opus 5 on EMSDialog (76.7\% vs.\ 82.8\%) and MediTOD (51.2\% vs.\ 53.6\%). It achieves more stable post-commit predictions compared with GPT-6 Sol (16.4\% vs.\ 18.1\% PC-Err) on EMSDialog, although GPT-6 Sol achieves higher earliness (Table~\ref{tab:main_results}). At matched dialogue progress, \textsc{DoC} achieves comparable accuracy to the in-context baselines on EMSDialog and MediTOD (Figure~\ref{fig:acc_vs_commit}).

\noindent\textbf{Deferral decisions better align with diagnostic errors.} \textsc{DoC} achieves the highest Acc@Def on EMSDialog and MediTOD (79.3\% and 61.8\%, Table~\ref{tab:main_results}), with RL adding 5.6 and 5.8 points over SFT. %
Its deferrals thus align more
closely with when its forced diagnosis would be wrong.

\subsection{RQ2: Ablation Studies}
\label{sec:rq2}

\begin{table}[t]
\centering
\small
\setlength{\tabcolsep}{1.5pt}
\renewcommand{\arraystretch}{1}
\begin{tabular*}{\columnwidth}{@{\extracolsep{\fill}}c ccc cc@{}}
\toprule
\multirow{2}{*}{$k$}
& \multicolumn{3}{c}{\textbf{Labeler $f_k$ (Val)}}
& \multicolumn{2}{c}{\textbf{SFT $\pi_{\mathrm{SFT}}^{(k)}$ (Test)}} \\
\cmidrule(lr){2-4}
\cmidrule(lr){5-6}
& Acc@1.0
& $\bar p_{\mathrm{val}}^{*(k)}$
& $\Delta\bar p_{\mathrm{val}}^*$
& Commit pos.
& Acc@Cmt \\
\midrule
Fixed & -- & -- & -- & 43.4 & 73.3 \\
\midrule
0 & 80.6 & 26.8 & --     & 29.7 & 73.3 \\
1 & 82.4 & 24.9 & $-1.9$ & 25.9 & 75.0 \\
2 & 85.2 & 23.5 & $-1.4$ & 23.9 & 75.0 \\
\textbf{3}
& \textbf{86.5}
& \textbf{22.5}
& $\mathbf{-1.0}$
& \textbf{23.2}
& \textbf{75.0} \\
\textcolor{gray}{4}
& \textcolor{gray}{85.8}
& \textcolor{gray}{23.0}
& \textcolor{gray}{$+0.5$}
& \textcolor{gray}{24.5}
& \textcolor{gray}{73.3} \\
\bottomrule
\end{tabular*}
\vspace{-0.5em}
\caption{EMSDialog boundary estimation (seed 42; values in \%).
Fixed: $t_i^*=0.2\,T_i$. Bold/gray: select/reject.}
\label{tab:rounds}
\vspace{-1em}
\end{table}

\begin{figure}[t]
    \centering
    \includegraphics[width=\columnwidth]
    {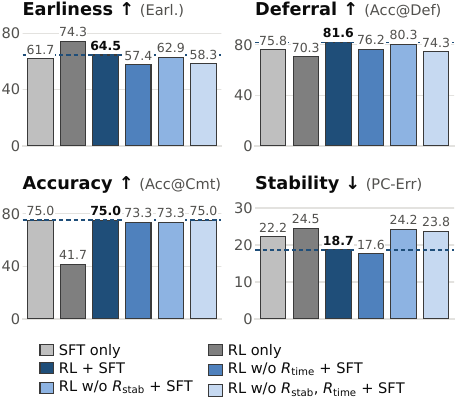}
    \vspace{-1.5em}
    \caption{
    Ablations of SFT and RL reward on EMSDialog (seed 42).
    Dashed lines indicate the full model.
    }
    \label{fig:ablation}
    \vspace{-1em}
\end{figure}

\noindent\textbf{Iterative boundary estimation enables earlier commitments without sacrificing accuracy.} Compared with the most accurate fixed-boundary baseline, which sets
the commitment boundary at 20\% of each dialogue ($t_i^*=0.2\,T_i$; Appendix~\ref{app:fixed_boundary}), the initial estimated boundaries yield earlier commitments (29.7\% vs.\ 43.4\% of dialogue progress) at the same Acc@Commit of 73.3\% (Table~\ref{tab:rounds}). Refining the boundaries further advances the SFT policy's mean commitment position to 23.2\% of dialogue progress while slightly improving Acc@Commit from 73.3\% to 75.0\%. The subsequent rejected round ($k=4$) moves boundaries later and lowers downstream policy accuracy, supporting a stopping criterion over assuming that more iterations always improve supervision.

\noindent\textbf{SFT Initialization and RL provide
complementary benefits.} Figure~\ref{fig:ablation} shows that RL only achieves high Earliness (74.3) but low Acc@Commit (41.7\%) and Acc@Def (70.3\%), indicating that early correct predictions do not necessarily imply reliable first commitments. SFT provides a more reliable starting point: on EMSDialog, adding RL preserves its Acc@Commit at 75.0\% while improving Earliness from 61.7 to 64.5 and reducing PC-Err from 22.2\% to 18.7\%. The combination thus improves timeliness and stability without sacrificing first-commit accuracy.

\noindent\textbf{The reward terms improve distinct aspects
of the prediction trajectory.} Removing the timeliness reward lowers Earliness (57.4 vs.\ 64.5), whereas removing the stability reward raises PC-Err (24.2\% vs.\ 18.7\%) (Figure~\ref{fig:ablation}). Removing both terms degrades both metrics, while Acc@Commit remains similar across these reward ablations. First-commit accuracy alone therefore misses the effects of the training objective, supporting explicit optimization of both timely correct predictions and sustained post-commit correctness.

\subsection{RQ3: Clinician Study}
\label{sec:rq3}

An EMT-certified clinician evaluated 60 EMSDialog test dialogues (seed 42) in two stages. First, with reference labels hidden, the clinician read each transcript incrementally, at the same prefixes as the models, and selected \textsc{Defer} or a diagnosis. Second, with reference labels visible, the clinician reviewed each \textsc{DoC} (SFT+RL) trajectory alongside the encounter audio and rated it as \textit{helpful}, \textit{neutral}, or \textit{harmful}, giving one primary reason per non-helpful
rating (Appendix~\ref{app:clinician_protocol}; Figures~\ref{fig:error-case-1}--\ref{fig:error-case-6}). Six of the final clinician diagnoses differed from the reference labels, possibly because of  clinical information that was not verbalized or details missing from the simulated scenarios.

\begin{figure}[t]
    \centering
    \includegraphics[width=\columnwidth]{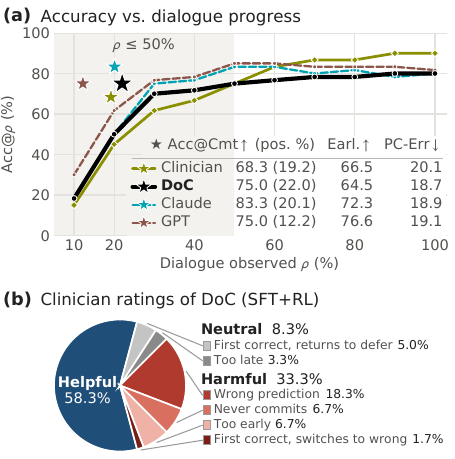}
    \vspace{-1em}
    \caption{Clinician study on the 60 EMSDialog test dialogues. \emph{Clinician}: the clinician's own diagnoses.}
    \label{fig:human_eval}
    \vspace{-1em}
\end{figure}

\noindent\textbf{DoC matches or exceeds the clinician early.} \textsc{DoC} is at least as accurate as the clinician through $\rho=50\%$ (70.0\% vs.\ 61.7\% at 30\%), but less accurate on the complete dialogue (80.0\% vs.\ 90.0\%). Its first commitments are more accurate (75.0\% vs.\ 68.3\%), with lower post-commit error (18.7\% vs.\ 20.1\%), though slightly later (22.0\% vs.\ 19.2\% of the dialogue). Claude Opus 5 and GPT-6 Sol also outperform the clinician early and reach correct diagnoses earlier than \textsc{DoC},
suggesting that model predictions may serve as early assistive signals. 

\noindent\textbf{Clinician ratings reflect both accuracy and commitment
behavior.} The clinician rated 35 of 60 trajectories (58.3\%) helpful, 5 (8.3\%) neutral, and 20 (33.3\%) harmful (Figure~\ref{fig:human_eval}b). Of the 25 non-helpful ratings (41.7\%) 11 cited wrong predictions (18.3\%), 6 commitment timing (10.0\%), 4 non-commitment (6.7\%), and 4 post-commit changes (6.7\%). Diagnostic errors are thus the most common
concern, while the remaining concerns involve the commitment behavior
targeted by our rewards. %
See Appendix~\ref{app:error_cases} for a complementary analysis. %

\section{Conclusion}

This paper introduced \textsc{DoC}, a framework that learns when to defer and when to commit in conversational diagnosis prediction. \textsc{DoC} derives supervision from iteratively estimated commitment boundaries and refines the policy with trajectory-level rewards for timeliness, first-commit accuracy, and post-commit stability. On EMSDialog and MediTOD, it attains the highest first-commit accuracy among post-trained methods and the lowest post-commit error rate overall. Ablations attribute earlier commitments to boundary estimation and complementary gains to supervised initialization and the reward terms. A clinician study shows that its early accuracy matches or exceeds the clinician's and that most trajectories are rated helpful.

\section*{Limitations}
First, the CDP task assumes a single ground-truth diagnosis per dialogue and treats diagnosis prediction as multiclass classification. This oversimplifies encounters involving multiple coexisting conditions or differential diagnoses that evolve as evidence accumulates. In such settings, a single diagnosis label may not represent the patient's conditions or the remaining diagnostic uncertainty. Extending the framework to predict multiple diagnoses or maintain a differential diagnosis, with potentially different commitment times per condition, is an important direction for future work.

Second, our evaluation elicits a decision every two utterances. We have not evaluated streaming audio input with fixed-duration windows (e.g., four-second intervals), which may split utterances and expose different amounts of evidence with each input update. Future work should assess whether the observed accuracy, timeliness, and stability transfer to this setting and also measure end-to-end latency.

Third, \textsc{DoC} uses only dialogue transcripts as input. EMS responders do not always verbalize what they observe or do: visual findings such as skin color, monitor readings, and performed interventions may go unspoken and are thus absent from the text. This may limit the accuracy and earliness achievable by any text-only predictor. Future works could incorporate multimodal inputs, such as vital-sign streams, ECG or video, to help the model commit earlier and more accurately.

Fourth, our data is limited in scale and realism. EMSDialog training
dialogues are synthetic, both test sets are small (60 and 69
dialogues), and six MediTOD test dialogues carry labels unseen in
training, so results may vary across splits and may not transfer to
other clinical settings. In addition, the early-evidence boundary
relies on utterance-level topic annotations that many clinical corpora lack; replacing them with automatic topic tagging is a natural extension.

Lastly, our clinician study involves a single EMT-certified clinician and 60 transcribed EMS simulations, so its comparisons are descriptive. Its ratings are post-hoc judgments of recorded dialogues rather than measurements of real-time assistance, and were made with reference labels visible, which may bias them. Whether model predictions improve clinical decision-making or patient outcomes remains unestablished. Our models are intended for research purposes and are not validated for clinical deployment. Prospective evaluation in clinical workflows is necessary to assess their effectiveness, safety, and influence on clinician behavior.

\bibliography{custom}

\begin{thebibliography}{29}
\providecommand{\natexlab}[1]{#1}

\bibitem[{{Anthropic}(2026)}]{anthropic2026opus5}
{Anthropic}. 2026.
\newblock \href
  {https://www-cdn.anthropic.com/c5fbac3f0b1280a933ebd26d3cb8bb9f5bdeaf48/Claude%20Opus%205%20System%20Card.pdf}
  {System card: {Claude Opus 5}}.
\newblock July 24, 2026.

\bibitem[{Bodenreider(2004)}]{bodenreider2004unified}
Olivier Bodenreider. 2004.
\newblock The unified medical language system (umls): integrating biomedical
  terminology.
\newblock \emph{Nucleic acids research}, 32(suppl\_1):D267--D270.

\bibitem[{Chang and
  Danescu-Niculescu-Mizil(2019)}]{chang-danescu-niculescu-mizil-2019-trouble}
Jonathan~P. Chang and Cristian Danescu-Niculescu-Mizil. 2019.
\newblock \href {https://doi.org/10.18653/v1/D19-1481} {Trouble on the horizon:
  Forecasting the derailment of online conversations as they develop}.
\newblock In \emph{Proceedings of the 2019 Conference on Empirical Methods in
  Natural Language Processing and the 9th International Joint Conference on
  Natural Language Processing (EMNLP-IJCNLP)}, pages 4743--4754, Hong Kong,
  China. Association for Computational Linguistics.

\bibitem[{Chen et~al.(2023)Chen, Yoon, Ebrahimi, Arik, Pfister, and
  Jha}]{chen-etal-2023-adaptation}
Jiefeng Chen, Jinsung Yoon, Sayna Ebrahimi, Sercan Arik, Tomas Pfister, and
  Somesh Jha. 2023.
\newblock \href {https://doi.org/10.18653/v1/2023.findings-emnlp.345}
  {Adaptation with self-evaluation to improve selective prediction in {LLM}s}.
\newblock In \emph{Findings of the Association for Computational Linguistics:
  EMNLP 2023}, pages 5190--5213, Singapore. Association for Computational
  Linguistics.

\bibitem[{Feng et~al.(2024)Feng, Shi, Wang, Ding, Balachandran, and
  Tsvetkov}]{feng2024don}
Shangbin Feng, Weijia Shi, Yike Wang, Wenxuan Ding, Vidhisha Balachandran, and
  Yulia Tsvetkov. 2024.
\newblock Don’t hallucinate, abstain: Identifying llm knowledge gaps via
  multi-llm collaboration.
\newblock In \emph{Proceedings of the 62nd Annual Meeting of the Association
  for Computational Linguistics (Volume 1: Long Papers)}, pages 14664--14690.

\bibitem[{Ge et~al.(2026)Ge, Murtaza, Cortez, and
  Alemzadeh}]{ge-etal-2026-emsdialog}
Xueren Ge, Sahil Murtaza, Anthony Cortez, and Homa Alemzadeh. 2026.
\newblock \href {https://doi.org/10.18653/v1/2026.findings-acl.1751}
  {{EMSD}ialog: Synthetic multi-person emergency medical service dialogue
  generation from electronic patient care reports via multi-{LLM} agents}.
\newblock In \emph{Findings of the {A}ssociation for {C}omputational
  {L}inguistics: {ACL} 2026}, pages 35081--35110, San Diego, California, United
  States. Association for Computational Linguistics.

\bibitem[{Gupta et~al.(2020)Gupta, Gupta, Biswas, and
  Dutta}]{gupta2020approaches}
Ashish Gupta, Hari~Prabhat Gupta, Bhaskar Biswas, and Tanima Dutta. 2020.
\newblock Approaches and applications of early classification of time series: A
  review.
\newblock \emph{IEEE Transactions on Artificial Intelligence}, 1(1):47--61.

\bibitem[{He et~al.(2020)He, Gu, Shen, and Ranzato}]{He2020Revisiting}
Junxian He, Jiatao Gu, Jiajun Shen, and Marc'Aurelio Ranzato. 2020.
\newblock \href {https://openreview.net/forum?id=SJgdnAVKDH} {Revisiting
  self-training for neural sequence generation}.
\newblock In \emph{International Conference on Learning Representations}.

\bibitem[{Kaushal et~al.(2023)Kaushal, Gupta, Upadhyay, and
  Faruqui}]{kaushal2023efficient}
Ayush Kaushal, Aditya Gupta, Shyam Upadhyay, and Manaal Faruqui. 2023.
\newblock Efficient encoders for streaming sequence tagging.
\newblock In \emph{Proceedings of the 17th Conference of the European Chapter
  of the Association for Computational Linguistics}, pages 418--429.

\bibitem[{Kementchedjhieva and
  S{\o}gaard(2021)}]{kementchedjhieva-sogaard-2021-dynamic}
Yova Kementchedjhieva and Anders S{\o}gaard. 2021.
\newblock \href {https://doi.org/10.18653/v1/2021.emnlp-main.624} {Dynamic
  forecasting of conversation derailment}.
\newblock In \emph{Proceedings of the 2021 Conference on Empirical Methods in
  Natural Language Processing}, pages 7915--7919, Online and Punta Cana,
  Dominican Republic. Association for Computational Linguistics.

\bibitem[{Kim et~al.(2021)Kim, Guo, Williams, Stankovic, and
  Alemzadeh}]{kim2021information}
Sion Kim, Weishi Guo, Ronald Williams, John Stankovic, and Homa Alemzadeh.
  2021.
\newblock Information extraction from patient care reports for intelligent
  emergency medical services.
\newblock In \emph{2021 IEEE/ACM Conference on Connected Health: Applications,
  Systems and Engineering Technologies (CHASE)}, pages 58--69. IEEE.

\bibitem[{Lai et~al.(2026)Lai, Liu, Wang, Ma, and Liu}]{lai2026doctorr}
Yunghwei Lai, Kaiming Liu, Ziyue Wang, Weizhi Ma, and Yang Liu. 2026.
\newblock \href {https://openreview.net/forum?id=vQGHTyL0Jw} {Doctor-r1:
  Mastering clinical inquiry with experiential agentic reinforcement learning}.
\newblock In \emph{The Fourteenth International Conference on Learning
  Representations}.

\bibitem[{Li et~al.(2024)Li, Balachandran, Feng, Ilgen, Pierson, Koh, and
  Tsvetkov}]{li2024mediq}
Shuyue~S Li, Vidhisha Balachandran, Shangbin Feng, Jonathan~S Ilgen, Emma
  Pierson, Pang~W Koh, and Yulia Tsvetkov. 2024.
\newblock Mediq: Question-asking llms and a benchmark for reliable interactive
  clinical reasoning.
\newblock \emph{Advances in Neural Information Processing Systems},
  37:28858--28888.

\bibitem[{Liu et~al.(2025)Liu, Qu, Schneider, Singh, and
  Kumar}]{liu2025cartteachingllmagents}
Grace Liu, Yuxiao Qu, Jeff Schneider, Aarti Singh, and Aviral Kumar. 2025.
\newblock \href {https://arxiv.org/abs/2510.08517} {Cart: Teaching llm agents
  to know when they know enough}.
\newblock \emph{Preprint}, arXiv:2510.08517.

\bibitem[{Liu et~al.(2022)Liu, Capurro, Nguyen, and
  Verspoor}]{liu-etal-2022-improving}
Jinghui Liu, Daniel Capurro, Anthony Nguyen, and Karin Verspoor. 2022.
\newblock \href {https://aclanthology.org/2022.alta-1.10/} {Improving
  text-based early prediction by distillation from privileged time-series
  text}.
\newblock In \emph{Proceedings of the 20th Annual Workshop of the Australasian
  Language Technology Association}, pages 73--83, Adelaide, Australia.
  Australasian Language Technology Association.

\bibitem[{Loshchilov and Hutter(2019)}]{loshchilov2018decoupled}
Ilya Loshchilov and Frank Hutter. 2019.
\newblock \href {https://openreview.net/forum?id=Bkg6RiCqY7} {Decoupled weight
  decay regularization}.
\newblock In \emph{International Conference on Learning Representations}.

\bibitem[{Machcha et~al.(2026)Machcha, Yerra, Gupta, Sahoo, Sultana, Yu, and
  Yao}]{machcha-etal-2026-knowing}
Sravanthi Machcha, Sushrita Yerra, Sahil Gupta, Aishwarya Sahoo, Sharmin
  Sultana, Hong Yu, and Zonghai Yao. 2026.
\newblock \href {https://doi.org/10.18653/v1/2026.eacl-long.291} {Knowing when
  to abstain: Medical {LLM}s under clinical uncertainty}.
\newblock In \emph{Proceedings of the 19th Conference of the {E}uropean Chapter
  of the {A}ssociation for {C}omputational {L}inguistics (Volume 1: Long
  Papers)}, pages 6153--6182, Rabat, Morocco. Association for Computational
  Linguistics.

\bibitem[{Madureira and Schlangen(2020)}]{madureira-schlangen-2020-incremental}
Brielen Madureira and David Schlangen. 2020.
\newblock \href {https://doi.org/10.18653/v1/2020.emnlp-main.26} {Incremental
  processing in the age of non-incremental encoders: An empirical assessment of
  bidirectional models for incremental {NLU}}.
\newblock In \emph{Proceedings of the 2020 Conference on Empirical Methods in
  Natural Language Processing (EMNLP)}, pages 357--374, Online. Association for
  Computational Linguistics.

\bibitem[{{OpenAI}(2026)}]{openai2026gpt6sol}
{OpenAI}. 2026.
\newblock \href {https://openai.com/index/introducing-gpt-6-sol-and-luna/}
  {Introducing {GPT-6} {Sol} and {Luna}}.
\newblock September 22, 2026.

\bibitem[{Saley et~al.(2024)Saley, Saha, Das, Raghu, and
  Mausam}]{saley-etal-2024-meditod}
Vishal~Vivek Saley, Goonjan Saha, Rocktim~Jyoti Das, Dinesh Raghu, and Mausam.
  2024.
\newblock \href {https://doi.org/10.18653/v1/2024.emnlp-main.936} {{M}edi{TOD}:
  An {E}nglish dialogue dataset for medical history taking with comprehensive
  annotations}.
\newblock In \emph{Proceedings of the 2024 Conference on Empirical Methods in
  Natural Language Processing}, pages 16843--16877, Miami, Florida, USA.
  Association for Computational Linguistics.

\bibitem[{Shao et~al.(2024)Shao, Wang, Zhu, Xu, Song, Bi, Zhang, Zhang, Li, Wu,
  and Guo}]{shao2024deepseekmathpushinglimitsmathematical}
Zhihong Shao, Peiyi Wang, Qihao Zhu, Runxin Xu, Junxiao Song, Xiao Bi, Haowei
  Zhang, Mingchuan Zhang, Y.~K. Li, Y.~Wu, and Daya Guo. 2024.
\newblock \href {https://arxiv.org/abs/2402.03300} {Deepseekmath: Pushing the
  limits of mathematical reasoning in open language models}.
\newblock \emph{Preprint}, arXiv:2402.03300.

\bibitem[{Shu et~al.(2019)Shu, Preum, Pitchford, Williams, Stankovic, and
  Alemzadeh}]{shu2019behavior}
Sile Shu, Sarah Preum, Haydon~M Pitchford, Ronald~D Williams, John Stankovic,
  and Homa Alemzadeh. 2019.
\newblock A behavior tree cognitive assistant system for emergency medical
  services.
\newblock In \emph{2019 IEEE/RSJ International Conference on Intelligent Robots
  and Systems (IROS)}, pages 6188--6195. IEEE.

\bibitem[{Tomani et~al.(2024)Tomani, Chaudhuri, Evtimov, Cremers, and
  Ibrahim}]{tomani2024uncertainty}
Christian Tomani, Kamalika Chaudhuri, Ivan Evtimov, Daniel Cremers, and Mark
  Ibrahim. 2024.
\newblock Uncertainty-based abstention in llms improves safety and reduces
  hallucinations.
\newblock \emph{arXiv preprint arXiv:2404.10960}.

\bibitem[{Wang et~al.(2024)Wang, Zi, Zhao, Deng, and Qin}]{wang-etal-2024-esdm}
Bichen Wang, Yuzhe Zi, Yanyan Zhao, Pengfei Deng, and Bing Qin. 2024.
\newblock \href {https://aclanthology.org/2024.lrec-main.556/} {{ESDM}: Early
  {S}ensing depression model in social media streams}.
\newblock In \emph{Proceedings of the 2024 Joint International Conference on
  Computational Linguistics, Language Resources and Evaluation (LREC-COLING
  2024)}, pages 6288--6298, Torino, Italia. ELRA and ICCL.

\bibitem[{Weerasinghe et~al.(2024)Weerasinghe, Janapati, Ge, Kim, Iyer,
  Stankovic, and Alemzadeh}]{10.1109/IoTDI61053.2024.00012}
Keshara Weerasinghe, Saahith Janapati, Xueren Ge, Sion Kim, Sneha Iyer, John~A.
  Stankovic, and Homa Alemzadeh. 2024.
\newblock \href {https://doi.org/10.1109/IoTDI61053.2024.00012} {Real-time
  multimodal cognitive assistant for emergency medical services}.
\newblock In \emph{2024 IEEE/ACM Ninth International Conference on
  Internet-of-Things Design and Implementation (IoTDI)}, page 85–96. IEEE
  Press.

\bibitem[{Yang et~al.(2025)Yang, Li, Yang, Zhang, Hui, Zheng, Yu, Gao, Huang,
  Lv, Zheng, Liu, Zhou, Huang, Hu, Ge, Wei, Lin, Tang, Yang, Tu, Zhang, Yang,
  Yang, Zhou, Zhou, Lin, Dang, Bao, Yang, Yu, Deng, Li, Xue, Li, Zhang, Wang,
  Zhu, Men, Gao, Liu, Luo, Li, Tang, Yin, Ren, Wang, Zhang, Ren, Fan, Su,
  Zhang, Zhang, Wan, Liu, Wang, Cui, Zhang, Zhou, and
  Qiu}]{yang2025qwen3technicalreport}
An~Yang, Anfeng Li, Baosong Yang, Beichen Zhang, Binyuan Hui, Bo~Zheng, Bowen
  Yu, Chang Gao, Chengen Huang, Chenxu Lv, Chujie Zheng, Dayiheng Liu, Fan
  Zhou, Fei Huang, Feng Hu, Hao Ge, Haoran Wei, Huan Lin, Jialong Tang, and 41
  others. 2025.
\newblock \href {https://arxiv.org/abs/2505.09388} {Qwen3 technical report}.
\newblock \emph{Preprint}, arXiv:2505.09388.

\bibitem[{Yuan and Singh(2023)}]{Yuan_Singh_2023}
Jiaqing Yuan and Munindar~P. Singh. 2023.
\newblock \href {https://doi.org/10.1609/icwsm.v17i1.22200} {Conversation
  modeling to predict derailment}.
\newblock \emph{Proceedings of the International AAAI Conference on Web and
  Social Media}, 17(1):926--935.

\bibitem[{Zhang et~al.(2024)Zhang, Diao, Lin, Fung, Lian, Wang, Chen, Ji, and
  Zhang}]{zhang-etal-2024-r}
Hanning Zhang, Shizhe Diao, Yong Lin, Yi~Fung, Qing Lian, Xingyao Wang, Yangyi
  Chen, Heng Ji, and Tong Zhang. 2024.
\newblock \href {https://doi.org/10.18653/v1/2024.naacl-long.394} {{R}-tuning:
  Instructing large language models to say `{I} don{'}t know'}.
\newblock In \emph{Proceedings of the 2024 Conference of the North American
  Chapter of the Association for Computational Linguistics: Human Language
  Technologies (Volume 1: Long Papers)}, pages 7113--7139, Mexico City, Mexico.
  Association for Computational Linguistics.

\bibitem[{Zhang et~al.(2025)Zhang, McKeown, and
  Muresan}]{zhang-etal-2025-forecasting}
Yunfan Zhang, Kathleen McKeown, and Smaranda Muresan. 2025.
\newblock \href {https://aclanthology.org/2025.inlg-main.40/} {Forecasting
  conversation derailments through generation}.
\newblock In \emph{Proceedings of the 18th International Natural Language
  Generation Conference}, pages 699--715, Hanoi, Vietnam. Association for
  Computational Linguistics.

\end{thebibliography}

\appendix

\section{Appendix}
\label{sec:appendix}

\subsection{Reward Illustration}
\label{app:reward_fig}

\begin{figure*}[t]
    \centering
    \includegraphics[width=\textwidth]{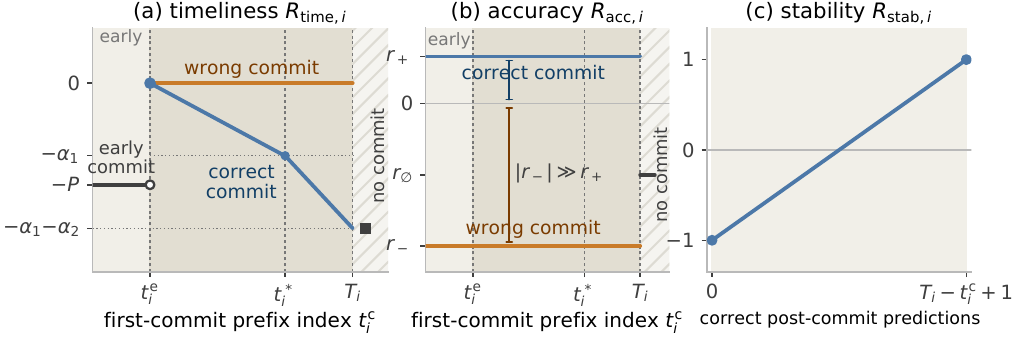}
    \vspace{-0.5em}
    \caption{Schematic illustration of the three reward terms: (a) timeliness, (b) first-commit accuracy, and (c) post-commit stability.}
    \label{fig:reward_appendix}
    \vspace{-1em}
\end{figure*}

Table~\ref{tab:reward_cases} defines the three reward terms, and Figure~\ref{fig:reward_appendix} visualizes their behavior. In panel (a), first commitments at or before $t_i^{\mathrm{e}}$ receive $-P$ regardless of correctness. After this boundary, correct commitments incur a piecewise linear delay penalty, while incorrect commitments receive zero timeliness reward. Non-commitment incurs the same delay penalty as a correct commitment at $T_i$. Panel (b) assigns accuracy rewards independently of commitment time, with $r_-<r_\varnothing<r_+$: an incorrect first commitment receives a lower accuracy reward than non-commitment. Panel (c) shows how the stability reward increases linearly with the proportion of correct post-commit predictions, from $-1$ when none are correct to $+1$ when all are correct. It equals zero when half are correct; a policy that never commits also receives zero.

\subsection{Conversation Zones}
\label{appendix:conv_zones}
We derive conversation zones from dataset-specific dialogue annotations. Below, we describe the annotation categories and their mapping to the early zone which is used to determine $t_i^{\mathrm{e}}$.
\paragraph{EMSDialog.}
The typical EMS dialogue topic flow includes \textit{Dispatch}, \textit{Introduction}, \textit{Chief Complaint}, \textit{Response Test}, \textit{HPI/Pain Assessment}, \textit{Primary Assessment},
\textit{Secondary Assessment}, \textit{Take Vitals},
\textit{Take Interventions}, \textit{Exit to Protocol}, \textit{Reassessment}, \textit{Transport}~\cite{kim2021information, ge-etal-2026-emsdialog}. These topics trace the progression from initial patient contact and complaint elicitation through assessment, intervention, and reassessment. We designate \textit{Dispatch}, \textit{Introduction} and \textit{Chief Complaint} as early-zone topics in EMS because they form the pre-assessment opening of an encounter. Information exchanged in this phase, such as the chief complaint, is often consistent with many conditions, (e.g., a confused or unresponsive patient may fall under hypoglycemia, stroke/TIA, opioid overdose, or seizure), so a diagnosis committed in this zone is often a guess even when correct.

\paragraph{MediTOD.}
Utterances are annotated using intents and slots. The intent categories are \textit{Inquire}, \textit{Inform}, \textit{Diagnose}, \textit{Salutations}, \textit{Chit-chat}, and \textit{Other}. The slot categories are \textit{Symptom}, \textit{Patient Medical History}, \textit{Family Medical History}, \textit{Habits}, \textit{Exposure}, \textit{Medication}, \textit{Medical Test}, \textit{Disease}, \textit{Travel Information}, \textit{Occupation},
\textit{Residence}, and \textit{Basic Information}. 
We designate opening \textit{Salutations} and all \textit{Symptom}-slot \textit{Inquire}/\textit{Inform} utterances as early-zone utterances. This heuristic reflects the ambiguity of presenting symptoms, which may be compatible with multiple diagnoses. The symptom-based criterion applies throughout the dialogue, so later matching utterances can extend the early zone.

\paragraph{Zone boundaries.}
For both datasets, $t_i^{\mathrm{e}}$ is the index of the last prefix ending in an early-zone utterance (Eq.~\eqref{eq:early_boundary}); we set $t_i^{\mathrm{e}}=0$ if no such prefix exists. We verified that $t_i^{\mathrm{e}}<T_i$ holds for every training and validation dialogue in both datasets, which guarantees $t_i^*>t_i^{\mathrm{e}}$ (including the fallback $t_i^*=T_i$) and thus a positive denominator $t_i^*-t_i^{\mathrm{e}}$ in Table~\ref{tab:reward_cases}. Averaging the relative position of $t_i^{\mathrm{e}}$ across dialogues, the early-evidence boundary occurs at 6.1\% and 13.5\% of dialogue progress in the EMSDialog and MediTOD training sets, respectively. The late zone begins at the first explicit human statement of the ground-truth diagnosis. This boundary is used only to evaluate lateness (Appendix~\ref{app:macro_f1_lateness_edit_overhead}).

\subsection{Data Preprocessing}
\label{app:data_prep}

\begin{table}[t]
\centering
\small
\setlength{\tabcolsep}{4pt}
\renewcommand{\arraystretch}{1.1}
\begin{tabular*}{\columnwidth}{@{\extracolsep{\fill}}l cc@{\hspace{10pt}}cc@{}}
\toprule
& \multicolumn{2}{c}{\textbf{EMSDialog}} & \multicolumn{2}{c}{\textbf{MediTOD}} \\
& \multicolumn{2}{c}{35 total labels} & \multicolumn{2}{c}{33 total labels} \\
\cmidrule(lr){2-3}\cmidrule(lr){4-5}
& Train & Test & Train & Test \\
\midrule
Dialogues       & 4{,}411 & 60    & 162  & 69 \\
Mean utterances & 114.3    & 117.0 & 95.7 & 95.4 \\
Observed labels & 35      & 9     & 27   & 19 \\
\bottomrule
\end{tabular*}
\caption{Dataset statistics before validation holdout. Mean utterances are measured per dialogue.}
\label{tab:data_stats}
\end{table}

\paragraph{Label masking.}
In both datasets and all experiments, label names mentioned in the dialogue are masked: EMS protocol names in EMSDialog and diagnosis names in MediTOD.

\subsubsection{EMSDialog}
\label{app:ems_prep}

\paragraph{Training dialogues.}
The 4{,}411 training dialogues are the synthetic multi-party EMS dialogues~\cite{ge-etal-2026-emsdialog}. Each dialogue is annotated with one diagnosis from the Regional EMS protocol set (43 distinct protocols). Several protocols are adult and pediatric versions of the same protocol (e.g.\ cardiac arrest, protocols 2-7 and 8-2; head injury, 4-6 and 10-4; seizure, 3-12 and 9-12); we merge these, which yields the 35 diagnosis labels used throughout. Every utterance carries a role (responder, patient, bystander) and a topic annotation (Dispatch, Introduction, Chief Complaint, Response Test, \ldots). Topics are used only to determine the early-evidence boundary $t_i^{\mathrm{e}}$ (Eq.~\eqref{eq:early_boundary}) and are never shown to the models. We hold out 20\% of the training dialogues for validation.

\paragraph{Test dialogues.}
The test set consists of 60 de-identified transcripts of EMS scenarios (speaker roles First Responders, Patient, Bystander), each labeled with a single protocol mapped to the same 35-label set.

\subsubsection{MediTOD}
\label{app:meditod_prep}

MediTOD~\citep{saley-etal-2024-meditod} is released as raw doctor--patient transcripts with utterance-level annotations (intents and medical slots such as \emph{Symptom}, \emph{Medication}, \emph{Disease}). The release we obtained contains 231 respiratory-specialty dialogues; it provides no train/validation/test identifiers, and the diagnosis is a free-text slot value rather than a label from a fixed set. We built the diagnosis-prediction corpus as follows.

\paragraph{Diagnosis labels.}
We use the primary diagnosis recorded in the \emph{Disease} slot as the gold label. The 231 dialogues contain 72 distinct diagnosis strings, reflecting variation in spelling and diagnostic granularity. We manually normalize and consolidate these strings into 33 task-specific labels, using the Unified Medical Language System (UMLS) as a terminology reference~\citep{bodenreider2004unified}. This process includes both lexical normalization and grouping related diagnoses into broader categories. For example, ten variants, including \emph{covid}, \emph{COVID}, \emph{Covid-19}, \emph{covid\_19}, and \emph{Coronavirus}, are mapped to COVID-19. Other task-specific groupings include \emph{strep throat}, \emph{viral pharyngitis}, and \emph{Tonsillitis} into Pharyngitis/Tonsillitis; \emph{viral infection}, \emph{common cold}, and \emph{upper respiratory tract infection} into Viral Upper Respiratory Infection; and \emph{acute copd exacerbation / infection}, \emph{chronic bronchitis}, and \emph{Emphysema} into COPD Exacerbation. The complete mapping will be released with our code.

\paragraph{Data Split.}
Because the release provides no split, we partition the 231 dialogues at random (seed 42) into 162 training and 69 test dialogues and hold out 20\% of the training dialogues for validation. Owing to the long tail, 6 of the 69 test dialogues carry a label absent from training (Measles, Spontaneous Rib Fracture, Haemophilus influenzae infection, Hypersensitivity Pneumonitis, Bronchiolitis, Obstructive Sleep Apnea); all models select from the full 33-label set.

\subsection{Additional Implementation Details}
\label{app:implementation}

\paragraph{Post-training setup.}
Labelers $f_k$, the \textsc{Defer}-aware SFT policy $\pi_{\mathrm{SFT}}$ and RL policy $\pi_{\mathrm{RL}}$ use full-parameter fine-tuning with AdamW (weight decay 0.01, warm-up ratio 0.05) in bf16. The effective batch size is 16 for EMSDialog and 8 for MediTOD, with a maximum input length of 8{,}192 tokens.

\paragraph{Labeler training.}
We sweep learning rates over $\{1,\dots,6\}\times10^{-5}$ using the validation set for selection. The initial labeler $f_0$ is trained for five epochs on complete training dialogues. Subsequent labelers ($f_k$)  are trained for five epochs on the diagnosis-labeled prefixes constructed in each iteration.

\paragraph{Boundary estimation.}
For each prefix $x_{i,t}$, we sample $M=8$ predictions with temperature 0.3, top-$p$ 0.8, and at most 48 new tokens. We use a reliability threshold of $\kappa=0.75$; if no prefix satisfies the boundary criterion, we set $t_i^*=T_i$. The mean normalized boundary $\bar p_{\mathrm{val}}^*$ is computed on the validation split. A round is accepted only if validation accuracy does not decrease and the mean normalized boundary advances by more than $\delta=0.005$ (0.5 percentage points of dialogue progress). The procedure stops at the first rejected round and retains the
training-set boundaries from the last accepted iteration.

\paragraph{\textsc{Defer}-aware SFT.}
We train the SFT policy on $\mathcal{D}_{\mathrm{SFT}}$ for five epochs. We sweep learning rates over $\{1,2,3,4,5,6\}\times10^{-5}$ and use the validation set for selection. The selected checkpoint uses a learning rate of $3\times10^{-5}$, and initializes RL.

\paragraph{Reinforcement learning.}
We sample $G=8$ trajectories per dialogue with temperature 1.0, top-$p$ 1.0, and at most 48 new tokens per decision. For each batch of four dialogues, we accumulate gradients over all sampled trajectories and take a single optimizer step, without reusing the trajectories. This limits repeated forward and backward computation over every decision prefix of the $G$ trajectories (tens of prefixes per dialogue, with prompts of up to
8{,}192 tokens). We sweep learning rates over $\{1,2,3,4,5,6\}\times10^{-5}$
and KL coefficients over $\{0,0.005\}$. We use trajectory-level outcome credit. The reported configuration uses a learning rate of $3\times10^{-5}$ and a KL coefficient $\beta=0.005$. The gradient norm is clipped at 1.0.
Group-relative advantages are computed as $(R_i^{(g)}-\bar R_i)/(\sigma_i+\eta)$, with $\eta=10^{-6}$. Groups with $\sigma_i<10^{-6}$ do not contribute to the policy-gradient term. We train for up to five epochs and evaluate every 100 updates, selecting the checkpoint with the highest validation reward. The reward coefficients in Section~\ref{sec:setup} are fixed a priori and shared across both datasets.

\paragraph{Inference.}
Evaluation uses the dialogue prefixes specified in Section~\ref{sec:setup}. We limit generation to 48 new tokens per decision. We disable thinking for all in-context baseline LLMs and use each model’s default sampling temperature.

\paragraph{Baseline implementation.}
We implement each baseline following its original paper and released implementation, retaining the baseline-specific training and inference settings, including decision thresholds, except for the CDP adaptations described below. All trained components use Qwen3-1.7B. Every baseline receives the same dialogue prefixes, sampled every two utterances (Section~\ref{sec:setup}), and produces an action at each prefix, including after its first commitment.

Static and Dynamic Train~\cite{ge-etal-2026-emsdialog} use Qwen3-1.7B as an encoder: the final-layer hidden state of the last input token is passed through dropout and a linear classifier over the label set, and the whole model is fine-tuned with cross-entropy on complete dialogues (Static) or on the last $5$ utterance-level prefixes of each training dialogue (Dynamic); for each prefix they commit to the arg-max label if its softmax probability exceeds 0.5 and defer otherwise.

For CaRT~\cite{liu2025cartteachingllmagents}, we fine-tune a Qwen3-1.7B termination model on counterfactual sufficient/insufficient prefix pairs built from the training data. It is queried on each prefix until it first outputs \textsc{Terminate}; thereafter, a diagnosis model fine-tuned on complete dialogues predicts from the current prefix. We use the SFT-based CaRT variant, as the original study shows that additional RL delays termination while yielding smaller gains over a fixed-budget baseline with matched mean termination length.

ESDM~\cite{wang-etal-2024-esdm} evaluates its stopping module and partial-information classifier independently for each prefix and commits to the classifier's arg-max label when the stopping probability is at least 0.5, deferring otherwise, with a forced commitment on the complete dialogue.

For ASPIRE~\cite{chen-etal-2023-adaptation}, a task LoRA adapter generates a diagnosis for each prefix and a separately trained self-evaluation LoRA adapter estimates the probability $p_{\mathrm{corr}}$ that it is correct; the model commits when $0.75\log p_{\mathrm{diag}}+0.25\log p_{\mathrm{corr}}\ge -2$ and defers otherwise, where $p_{\mathrm{diag}}$ is the length-normalized likelihood of the generated diagnosis.

R-Tuning~\cite{zhang-etal-2024-r} is fine-tuned to follow each diagnosis with ``Are you sure you accurately answered the question based on your internal knowledge? I am sure/unsure'', where the label records whether the model predicted that training prefix correctly; for each prefix it generates a diagnosis and commits to it if $P(\textit{sure})\ge 0.5$ (normalized over \textit{sure} and \textit{unsure}), and defers otherwise.

\subsection{Evaluation Metric Details}
\label{app:evaluation_metrics}

For test dialogue $i$, let $x_{i,1},\ldots,x_{i,T_i}$ denote its evaluated prefixes and $y_i$ its ground-truth diagnosis. For each prefix $x_{i,t}$, the model produces an action $a_{i,t}\in\mathcal{Y}\cup\{\textsc{Defer}\}$. A commitment occurs whenever the model outputs a diagnosis rather than \textsc{Defer}. We retain the complete action trajectory, including predictions after the first commitment.

\paragraph{Accuracy at commitment (Acc@Commit ($\uparrow$)).}
Let $t_i^{\mathrm{c}}$ denote the index of the first prefix for which the model outputs a diagnosis (Eq.~\eqref{eq:first_commit}):
\[
t_i^{\mathrm{c}} = \min\{t : a_{i,t} \neq \textsc{Defer}\},
\]
with $t_i^{\mathrm{c}}=T_i+1$ if the model never commits. A dialogue is counted as correct if $t_i^{\mathrm{c}}\leq T_i$ and $a_{i,t_i^{\mathrm{c}}}=y_i$. Dialogues in which the model never commits are counted as incorrect. Acc@Commit is the proportion of all dialogues in which
the first committed diagnosis is correct. Because commitment timing is determined by the model, this metric evaluates predictions at potentially different stages of dialogue progress.

\paragraph{Acc@$\rho$ ($\uparrow$).} Acc@$\rho$ measures diagnosis accuracy at a fixed dialogue
fraction $\rho \in (0,1]$~\citep{liu2025cartteachingllmagents}. For each dialogue, we multiply its total utterance count by $\rho$, round up, and select the longest evaluated prefix ending at or before this position. Acc@$\rho$ is the proportion of all dialogues whose recorded action at the selected prefix matches the ground-truth label; incorrect diagnoses and \textsc{Defer}
are counted as incorrect. Acc@1.0 evaluates the action on the complete dialogue.

\paragraph{Earliness ($\uparrow$).}
We measure first-correct earliness~\citep{gupta2020approaches}.
Let
\[
t_i^{\mathrm{fc}} = \min\{t : a_{i,t} = y_i\}
\]
denote the index of the first prefix yielding the correct diagnosis. The dialogue-level score is
\[
E_i =
\begin{cases}
1 - t_i^{\mathrm{fc}}/T_i, & \text{if a correct prediction occurs},\\
0, & \text{otherwise}.
\end{cases}
\]
We report the average of $E_i$ across dialogues. Higher scores indicate earlier correct predictions. This metric depends on the first correct prediction, regardless of whether subsequent predictions remain correct.

\paragraph{Post-commit error rate (PC-Err ($\downarrow$)).}
PC-Err is the post-commit analogue of streaming prediction
error~\citep{kaushal2023efficient}.
It measures the proportion of decisions from the first commitment onward (inclusive) at which the model fails to output the ground-truth diagnosis. For dialogue $i$ with $t_i^{\mathrm{c}}\leq T_i$,
\[
\mathrm{PC\text{-}Err}_i
=
\frac{\sum_{t=t_i^{\mathrm{c}}}^{T_i} \mathds{1} [a_{i,t} \neq y_i]}
     {T_i-t_i^{\mathrm{c}}+1}.
\]
Both incorrect diagnoses and returns to \textsc{Defer} count as errors. PC-Err includes the first commitment and is averaged over dialogues. Never-committing dialogues are excluded from PC-Err but counted as incorrect under Acc@Commit. For a dialogue that commits only at the final decision, PC-Err is zero for a correct commitment and one otherwise.

\paragraph{Deferral accuracy (Acc@Def ($\uparrow$)).}
Following~\citet{feng2024don}, we adapt abstain accuracy to prefix-level deferral decisions. For each prefix $x_{i,t}$, we obtain a forced diagnosis $\tilde{a}_{i,t}$ by re-decoding the same prompt with \textsc{Defer} disabled (See Appendix~\ref{app:prompt}). These predictions define the reference decisions and do not replace the original model outputs.

Taking deferral as the positive action, we define four counts
for each dialogue $i$:
\begin{itemize}[nosep, leftmargin=*]
    \item $\mathrm{TP}_i$: the model defers and the forced diagnosis is incorrect.
    \item $\mathrm{FP}_i$: the model defers but the forced diagnosis is correct.
    \item $\mathrm{FN}_i$: the model commits but the forced diagnosis is incorrect.
    \item $\mathrm{TN}_i$: the model commits and the forced diagnosis is correct.
\end{itemize}
Deferral accuracy is
\[
\mathrm{Acc@Def}_i
=
\frac{\mathrm{TP}_i+\mathrm{TN}_i}
{\mathrm{TP}_i+\mathrm{FP}_i+\mathrm{FN}_i+\mathrm{TN}_i}.
\]
We report the mean Acc@Def across dialogues, counting all decisions, including those after the first commitment.

\begin{table*}[t]
\centering
\small
\setlength{\tabcolsep}{4pt}
\renewcommand{\arraystretch}{1.1}
\resizebox{\textwidth}{!}{%
\begin{tabular}{@{}l*{8}{c}@{}}
\toprule
\multirow{2}{*}{\textbf{Method}}
& \multicolumn{4}{c}{\textbf{EMSDialog}}
& \multicolumn{4}{c}{\textbf{MediTOD}} \\
\cmidrule(lr){2-5}
\cmidrule(lr){6-9}
& \textbf{MaF@Cmt}$\uparrow$
& \textbf{MaF@1.0}$\uparrow$
& \textbf{Lateness}$\downarrow$
& \textbf{EO}$\downarrow$
& \textbf{MaF@Cmt}$\uparrow$
& \textbf{MaF@1.0}$\uparrow$
& \textbf{Lateness}$\downarrow$
& \textbf{EO}$\downarrow$ \\
\midrule
\multicolumn{9}{@{}l}{\textit{In-context}} \\

Qwen3-1.7B
& 4.5{\scriptsize$\pm$2.2}
& 1.6{\scriptsize$\pm$0.7}
& 52.8{\scriptsize$\pm$4.2}
& 98.9{\scriptsize$\pm$1.1}
& 13.5{\scriptsize$\pm$3.9}
& 15.0{\scriptsize$\pm$3.2}
& 51.2{\scriptsize$\pm$3.0}
& 90.5{\scriptsize$\pm$4.0} \\

Qwen3-4B
& 42.8{\scriptsize$\pm$5.9}
& 58.6{\scriptsize$\pm$4.4}
& 4.4{\scriptsize$\pm$2.5}
& 47.0{\scriptsize$\pm$4.2}
& 12.0{\scriptsize$\pm$3.3}
& 35.5{\scriptsize$\pm$3.7}
& 4.8{\scriptsize$\pm$2.2}
& 80.2{\scriptsize$\pm$4.1} \\

Doctor-R1-8B
& 36.3{\scriptsize$\pm$5.3}
& 45.5{\scriptsize$\pm$5.0}
& 9.4{\scriptsize$\pm$2.5}
& 49.5{\scriptsize$\pm$3.3}
& 22.6{\scriptsize$\pm$5.5}
& 43.4{\scriptsize$\pm$6.8}
& 4.8{\scriptsize$\pm$0.8}
& 82.3{\scriptsize$\pm$3.0} \\

Claude Opus 5
& 72.1{\scriptsize$\pm$3.5}
& 59.0{\scriptsize$\pm$2.6}
& 2.8{\scriptsize$\pm$1.0}
& 33.7{\scriptsize$\pm$0.7}
& 42.9{\scriptsize$\pm$3.1}
& 48.1{\scriptsize$\pm$3.2}
& \textbf{0.0}{\scriptsize$\pm$0.0}
& 68.2{\scriptsize$\pm$0.4} \\

GPT-6 Sol
& \textbf{74.7}{\scriptsize$\pm$3.7}
& \textbf{65.7}{\scriptsize$\pm$4.3}
& \textbf{1.1}{\scriptsize$\pm$1.0}
& 38.5{\scriptsize$\pm$1.1}
& \textbf{44.5}{\scriptsize$\pm$3.3}
& 43.2{\scriptsize$\pm$2.4}
& 1.9{\scriptsize$\pm$0.8}
& 74.2{\scriptsize$\pm$1.2} \\

\midrule
\multicolumn{9}{@{}l}{
\textit{Post-trained (Qwen3-1.7B)}} \\

ASPIRE
& 40.4{\scriptsize$\pm$6.1}
& 57.1{\scriptsize$\pm$6.3}
& 11.1{\scriptsize$\pm$2.5}
& 36.5{\scriptsize$\pm$1.1}
& 15.6{\scriptsize$\pm$2.3}
& 24.5{\scriptsize$\pm$5.7}
& 9.7{\scriptsize$\pm$1.7}
& 80.7{\scriptsize$\pm$2.5} \\

ESDM
& 39.4{\scriptsize$\pm$4.2}
& 34.8{\scriptsize$\pm$5.6}
& 9.4{\scriptsize$\pm$2.5}
& 31.7{\scriptsize$\pm$2.0}
& 12.7{\scriptsize$\pm$2.7}
& 15.4{\scriptsize$\pm$3.3}
& 7.7{\scriptsize$\pm$0.8}
& 65.2{\scriptsize$\pm$1.7} \\

R-Tuning
& 49.3{\scriptsize$\pm$6.0}
& 61.7{\scriptsize$\pm$7.2}
& 6.7{\scriptsize$\pm$1.7}
& 31.7{\scriptsize$\pm$2.7}
& 23.8{\scriptsize$\pm$5.7}
& 24.3{\scriptsize$\pm$5.8}
& 16.9{\scriptsize$\pm$4.4}
& 86.3{\scriptsize$\pm$1.3} \\

CaRT
& 40.0{\scriptsize$\pm$4.7}
& 54.2{\scriptsize$\pm$5.7}
& 10.6{\scriptsize$\pm$1.0}
& 29.4{\scriptsize$\pm$1.9}
& 10.5{\scriptsize$\pm$3.2}
& \textbf{\underline{49.7}}{\scriptsize$\pm$3.0}
& \underline{1.9}{\scriptsize$\pm$1.7}
& 70.2{\scriptsize$\pm$0.6} \\

Static Train
& 34.1{\scriptsize$\pm$5.0}
& 49.9{\scriptsize$\pm$6.5}
& 6.1{\scriptsize$\pm$1.9}
& 35.2{\scriptsize$\pm$1.8}
& 8.5{\scriptsize$\pm$3.4}
& 24.9{\scriptsize$\pm$4.6}
& 3.9{\scriptsize$\pm$1.7}
& 95.7{\scriptsize$\pm$1.5} \\

Dynamic Train
& 35.9{\scriptsize$\pm$4.3}
& 45.0{\scriptsize$\pm$6.2}
& \underline{3.9}{\scriptsize$\pm$1.0}
& 42.1{\scriptsize$\pm$1.5}
& 11.1{\scriptsize$\pm$3.3}
& 26.7{\scriptsize$\pm$5.9}
& 3.4{\scriptsize$\pm$0.8}
& 69.8{\scriptsize$\pm$2.0} \\

\rowcolor{gray!12}
\textbf{SFT (DoC)}
& 55.9{\scriptsize$\pm$3.8}
& 61.5{\scriptsize$\pm$3.5}
& 8.3{\scriptsize$\pm$1.7}
& 23.8{\scriptsize$\pm$1.5}
& 29.0{\scriptsize$\pm$3.0}
& 48.1{\scriptsize$\pm$2.3}
& 5.8{\scriptsize$\pm$1.4}
& 44.6{\scriptsize$\pm$1.8} \\

\rowcolor{gray!12}
\textbf{SFT + RL (DoC)}
& \underline{60.8}{\scriptsize$\pm$4.1}
& \underline{62.2}{\scriptsize$\pm$4.6}
& 7.2{\scriptsize$\pm$2.5}
& \textbf{\underline{20.5}}{\scriptsize$\pm$1.6}
& \underline{31.5}{\scriptsize$\pm$5.1}
& \textbf{\underline{49.7}}{\scriptsize$\pm$3.5}
& 3.4{\scriptsize$\pm$2.2}
& \textbf{\underline{42.6}}{\scriptsize$\pm$1.8} \\

\bottomrule
\end{tabular}%
}
\caption{
Additional evaluation metrics, complementing
Table~\ref{tab:main_results}.
MaF denotes macro-F1, and EO denotes edit overheads.
Results are mean $\pm$ sample standard deviation over three runs.
\textbf{Bold} indicates the best mean overall;
\underline{underlining} indicates the best among evaluated
open-weight models.
}
\label{tab:appendix_metrics}
\end{table*}

\subsection{Macro-F1, Lateness, and Edit Overheads}
\label{app:macro_f1_lateness_edit_overhead}

We complement the main results with diagnosis macro-F1 (MaF), lateness, and edit overheads (EO).
MaF@Cmt and MaF@1.0 evaluate diagnosis performance at first commitment and full dialogue progress, respectively, assigning equal weight to each diagnosis class. Lateness measures whether the model first commits after the human has explicitly stated the ground-truth diagnosis, also counting never-committing dialogues as late. EO evaluates post-commit prediction revisions while allowing one necessary correction from an initially incorrect diagnosis to the ground truth.

\paragraph{Macro-F1 ($\uparrow$).}
Let $\mathcal{Y}_{\mathrm{test}}$ denote the set of diagnosis labels appearing in the ground-truth annotations of the test set: 9 labels for EMSDialog and 19 for MediTOD.
We compute
\[
\mathrm{MaF}
=
\frac{1}{|\mathcal{Y}_{\mathrm{test}}|}
\sum_{c\in\mathcal{Y}_{\mathrm{test}}}
\frac{2\,\mathrm{TP}_c}
     {2\,\mathrm{TP}_c+\mathrm{FP}_c+\mathrm{FN}_c},
\]
where $\mathrm{TP}_c$, $\mathrm{FP}_c$, and $\mathrm{FN}_c$ are the true-positive, false-positive, and false-negative counts for class $c$ at the evaluated decision point. MaF@Cmt uses the first committed diagnosis, whereas MaF@1.0 uses the final decision. Labels absent from the test-set ground truth are excluded from the macro average, although models still predict from the full 35-label or 33-label set.

\begin{table}[t]
\centering
\small
\setlength{\tabcolsep}{4pt}
\renewcommand{\arraystretch}{1.05}
\begin{tabular*}{\columnwidth}{@{\extracolsep{\fill}}lcw{c}{46pt}w{c}{46pt}@{}}
\toprule
& & \multicolumn{2}{c}{\makecell{\textbf{First human diagnosis}\\\textbf{(\% of dialogue)}}} \\
\cmidrule(lr){3-4}
\textbf{Dataset}
& \textbf{Test Size}
& \textbf{Mean}
& \textbf{Median} \\
\midrule
EMSDialog & 60 & 61.8 & 62.1 \\
MediTOD   & 69 & 97.6 & 98.4 \\
\bottomrule
\end{tabular*}
\caption[Position of the first human diagnosis statement]{
Position of the first explicit human protocol or diagnosis statement (percentage of the dialogue's total utterance count)
}
\label{tab:human_timing}
\end{table}

\paragraph{Lateness ($\downarrow$).}
For dialogue $i$, the reference point is the utterance in which the EMS crew or physician first explicitly states the ground-truth protocol or diagnosis; from this utterance on, a model prediction is late (start of the late zone in Figure~\ref{fig:pipeline}). In EMSDialog test data, we manually identify this statement from explicit diagnostic wording corresponding to a protocol label, such as ``you might be having a stroke'' or ``we're dealing with an opioid overdose.'' In the MediTOD test set, we use the first physician utterance whose \textit{Disease} slot corresponds to the ground-truth diagnosis.

Let $t_i^{\ell}$ index the longest evaluated prefix ending strictly before the reference utterance; if no such statement is recorded, we set $t_i^{\ell}=T_i$. A dialogue is late if the model first commits after $t_i^{\ell}$ (Eq.~\eqref{eq:first_commit}):
\[
L_i =
\mathds{1}\!\left[
t_i^{\mathrm{c}}>t_i^{\ell}
\right].
\]
Since $t_i^{\mathrm{c}}=T_i+1$ when the model never commits, never-committing dialogues count as late. A dialogue without a human diagnostic statement is late only if the model never commits.
Lateness is the mean of $L_i$ across dialogues, reported as a percentage; lower is better. It does not consider whether the first commitment is correct.

Table~\ref{tab:human_timing} summarizes the positions of observed human diagnostic statements.
These occur earlier in EMSDialog than in MediTOD: the mean positions are 61.8\% and 97.6\% of dialogue progress, respectively. This difference reflects the dialogue structure: in EMSDialog, the EMS crew usually states the protocol before patient transport and hospital handover, whereas MediTOD dialogues typically end with the physician summarizing the diagnosis. Thus, the observed reference points correspond to different stages of the conversation across datasets.

\paragraph{Edit Overheads ($\downarrow$).}
Following EMSDialog~\cite{ge-etal-2026-emsdialog}, we use Edit Overheads (EO) to quantify the instability of a model's committed predictions after its first commit. Let the post-commit sequence of predicted labels be $(a_{i,t_i^{\mathrm{c}}},\ldots,a_{i,T_i})$, where $a_{i,t_i^{\mathrm{c}}}$ is the first committed label and $y_i$ is the ground-truth label.
The number of post-commit action changes, including transitions to and from \textsc{Defer}, is
\[
C_i =
\sum_{t=t_i^{\mathrm{c}}+1}^{T_i}
\mathds{1}[a_{i,t}\ne a_{i,t-1}].
\]
Under their definition, a single correction is \emph{necessary}, $\mathrm{Necessary}_i=1$, if
$(a_{i,t_i^{\mathrm{c}}}\ne y_i)$ and $\exists\,t\in\{t_i^{\mathrm{c}},\ldots,T_i\}$ such that $a_{i,t}=y_i$; otherwise, $\mathrm{Necessary}_i=0$.
EO is computed as:
\begin{itemize}
    \item If $C_i=0$,
    $\mathrm{EO}_i=
    \mathds{1}[a_{i,t_i^{\mathrm{c}}}\ne y_i]$.

    \item If $C_i>0$, then
    \[
    \mathrm{EO}_i =
    \frac{C_i-\mathrm{Necessary}_i}{C_i}.
    \]
\end{itemize}
We average EO over dialogues with at least one commitment, excluding never-committing dialogues as in PC-Err. EO equals $0$ when the model stays correct without flipping, or makes only the single necessary correction from an initially incorrect label to the ground truth; EO approaches $1$ when the model frequently oscillates among labels or never reaches $y_i$.

\medskip
\noindent\textbf{RL improves macro-F1 on both datasets.} Table~\ref{tab:appendix_metrics} shows that adding RL increases MaF@Cmt from 55.9 to 60.8 on EMSDialog and from 29.0 to 31.5 on MediTOD, achieving the highest mean scores among post-trained methods on both datasets. The gains also extend to full-dialogue prediction: MaF@1.0 rises from 61.5 to 62.2 and from 48.1
to 49.7, respectively. These results indicate that the benefits of RL extend beyond overall accuracy to evaluation that weights diagnosis classes equally. However, both proprietary models retain higher MaF@Cmt, leaving a gap in class-balanced first-commit performance.

\medskip
\noindent\textbf{RL lowers lateness on both datasets.} Adding RL lowers lateness from 8.3\% to 7.2\% on EMSDialog and from 5.8\% to 3.4\% on MediTOD (Table~\ref{tab:appendix_metrics}).
The mean reduction on EMSDialog is modest, at 1.1 percentage points. Several post-trained baselines are late less often: Dynamic Train (3.9\%), Static Train (6.1\%), and R-Tuning (6.7\%) on EMSDialog, and CaRT (1.9\%) on MediTOD, where Dynamic Train ties \textsc{DoC} at 3.4\%. Each of these baselines has lower Acc@Commit than \textsc{DoC} (Table~\ref{tab:main_results}), so their lower lateness does not come with more accurate first commitments. GPT-6 Sol and Claude Opus 5 achieve the lowest lateness, at 1.1\% on EMSDialog and 0.0\% on MediTOD, respectively.

\medskip
\noindent\textbf{RL lowers edit overheads on both datasets.} Adding RL reduces EO from 23.8\% to 20.5\% on EMSDialog and from 44.6\% to 42.6\% on MediTOD, yielding the lowest mean EO among all evaluated methods on both datasets (Table~\ref{tab:appendix_metrics}). These reductions complement the lower PC-Err reported in Table~\ref{tab:main_results}: PC-Err measures the proportion of incorrect post-commit decisions, whereas EO evaluates prediction revisions while allowing one necessary correction. Together, these results support improved post-commit performance in terms of sustained correctness and edit overheads.

\begin{table}[t]
\centering\small
\setlength{\tabcolsep}{4pt}
\begin{tabular}{@{}c cc cc c@{}}
\toprule
$\phi$ & Commit pos. & Acc@Cmt & Acc@1.0 & Earl. & PC-Err \\
\midrule
0.10 & 22.1 & 65.0 & 68.3 & 56.2 & 32.7 \\
0.15 & 33.6 & 71.7 & 65.0 & 49.7 & 28.7 \\
0.20 & 43.4 & 73.3 & 70.0 & 42.1 & 24.5 \\
0.25 & 53.4 & 68.3 & 65.0 & 32.3 & 29.6 \\
0.30 & 57.7 & 53.3 & 50.0 & 30.3 & 32.2 \\
0.40 & 66.6 & 31.7 & 30.0 & 19.0 & 46.6 \\
0.50 & 72.8 & 26.7 & 20.0 & 15.4 & 51.6 \\
\midrule
Est. & 23.2 & 75.0 & 76.7 & 61.7 & 22.2 \\  %
\bottomrule
\end{tabular}

\caption[Fixed-fraction boundary control]{Fixed-fraction boundary control on EMSDialog (seed 42; metrics in \%).
Every training dialogue receives $t_i^*=\phi\,T_i$.
Est.: estimated boundaries ($k{=}3$, Table~\ref{tab:rounds}). Commit pos.: mean first-commit position $t_i^{\mathrm{c}}/T_i$.}
\label{tab:fixed_sweep}
\end{table}

\subsection{Fixed-Fraction Boundary Control}
\label{app:fixed_boundary}
To test whether dialogue-specific boundaries matter, we train the same \textsc{Defer}-aware SFT policy on boundaries that ignore dialogue content: every training dialogue receives $t_i^*=\phi\,T_i$ for a single fraction $\phi\in\{0.10,0.15,0.20,0.25,0.30,0.40,0.50\}$, kept after $t_i^{\mathrm{e}}$ as in Section~\ref{sec:method}. Only $t_i^*$ differs from the policies in Table~\ref{tab:rounds}; labels, training recipe, and evaluation are identical. Table~\ref{tab:fixed_sweep} reports results on the EMSDialog test set (seed 42).

Every fixed fraction yields lower Acc@Commit than the estimated boundaries (at most 73.3\% vs.\ 75.0\%). The only fraction that commits about as early ($\phi=0.10$, 22.1\% vs.\ 23.2\%) loses 10 points of Acc@Commit, and the most accurate one ($\phi=0.20$) commits about 20 points later (43.4\% vs.\ 23.2\%). From $\phi=0.30$ onward, Acc@Commit and Acc@1.0 fall together, to 26.7\% and 20.0\% at $\phi=0.50$, largely because the policy increasingly never commits: as a larger share of each training dialogue is labeled \textsc{Defer}, the policy learns to defer even on the complete dialogue. A fixed fraction therefore trades first-commit accuracy against delay, whereas the estimated boundaries improve both.

\subsection{Error Analysis}
\label{app:error_cases}

\begin{figure}[t]
  \centering
  \includegraphics[width=\columnwidth]{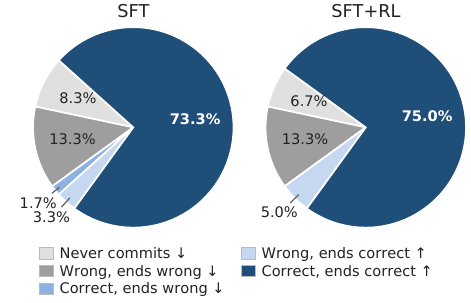}
  \caption{Error types on EMSDialog for SFT and SFT+RL (seed 42). $\downarrow$ / $\uparrow$: lower/higher is better.}
  \label{fig:traj_types}
\end{figure}

\paragraph{Changes in prediction trajectories.}
We categorize EMSDialog test trajectories by whether the model commits and by the correctness of its first committed diagnosis and final prediction: (i) never commits; (ii) initially incorrect and finally incorrect; (iii) initially incorrect but finally correct; (iv) initially correct but finally incorrect; and (v) initially correct and finally correct. For committed trajectories, a final \textsc{Defer} or invalid output counts as incorrect. These categories are mutually exclusive and describe first-to-final outcomes rather than intermediate prediction changes.

Figure~\ref{fig:traj_types} shows differences despite unchanged aggregate Acc@Commit. With RL, the proportion of trajectories that are initially correct but finally incorrect decreases from 1.7\% to 0\%, while the proportion correct at both points increases from 73.3\% to 75.0\%. Initially incorrect trajectories that end correctly increase from 3.3\% to 5.0\%, while those incorrect at both points remain at 13.3\%. Non-commitment decreases from 8.3\% to 6.7\%. These results associate RL with more correct final predictions and less non-commitment, distinctions that Acc@Commit alone does not capture.

\paragraph{Deferral decision breakdown.}
We complement the trajectory-level analysis with a decision-level view of deferral. Following Acc@Def (Appendix~\ref{app:evaluation_metrics}), each original action is compared with the correctness of the same model's forced diagnosis at the same prefix. Taking deferral as the positive action, TP denotes deferring when the forced diagnosis is incorrect, FP deferring when it is correct, FN committing when it is incorrect, and TN committing when it is correct. Table~\ref{tab:deferral_breakdown} pools these counts over all 3,524 decisions of the EMSDialog test set.

\begin{table}[t]
\centering
\small
\setlength{\tabcolsep}{4pt}
\begin{tabular*}{\columnwidth}{@{\extracolsep{\fill}}lrrrrcc@{}}
\toprule
\textbf{Model} & TP & FP & FN & TN & Prec. & Rec. \\
\midrule
DoC (SFT)     & 469 & 136 &  950 & 1969 & 77.5 & 33.1 \\
DoC (SFT+RL)  & 504 & 104 &  948 & 1968 & 82.9 & 34.7 \\
Claude Opus 5 & 413 & 191 &  859 & 2061 & 68.4 & 32.5 \\
GPT-6 Sol     & 290 &  81 &  977 & 2176 & 78.2 & 22.9 \\
\bottomrule
\end{tabular*}
\caption[Decision-level deferral breakdown]{
Decision-level deferral breakdown on the EMSDialog test set (seed 42), pooled over all 3,524 decisions per model. Positives are \textsc{Defer} actions; reference positives are incorrect forced diagnoses. Prec.\ $=\mathrm{TP}/(\mathrm{TP}+\mathrm{FP})$ and Rec.\ $=\mathrm{TP}/(\mathrm{TP}+\mathrm{FN})$.
}
\label{tab:deferral_breakdown}
\end{table}

\textsc{DoC} defers selectively but incompletely. Its deferral precision, the share of deferrals that coincide with an incorrect forced diagnosis, is 82.9\%, compared with 68.4\% for Claude Opus 5 and 78.2\% for GPT-6 Sol, at a deferral rate comparable to Claude Opus 5 (17.3\% vs.\ 17.1\% of decisions). Yet its deferral recall, the share of incorrect forced diagnoses on which it defers, is only 34.7\%, slightly above Claude Opus 5 (32.5\%); GPT-6 Sol defers less often (10.5\% of decisions) and reaches 22.9\%. RL improves both over SFT, from 77.5\% to 82.9\% precision and from 33.1\% to 34.7\% recall. It reduces pooled deferral errors (FP+FN) from 1,086 to 1,052, mostly by cutting unnecessary deferrals from 136 to 104, which is the decision-level counterpart of the Acc@Def gain in Table~\ref{tab:main_results}.

Two caveats apply. First, the reference positives are model-specific: the forced diagnosis is incorrect on 41.2\% of \textsc{DoC}'s decisions but on 36.1\% and 36.0\% of Claude Opus 5's and GPT-6 Sol's, so recall is not strictly comparable across models, and a correct forced diagnosis does not by itself establish that commitment was warranted. Second, these counts pool all decisions, so longer dialogues carry more weight.

\subsection{Clinician Study Protocol and Case Studies}
\label{app:clinician_protocol}

An EMT-certified clinician evaluated the 60 real EMSDialog test dialogues in two stages.

\paragraph{Stage 1: independent incremental diagnosis.}
For each dialogue prefix, the clinician selected one of the 35 diagnosis labels or \textsc{Defer} (Figure~\ref{fig:study-ui-stage1}). Model outputs and reference labels were hidden, and diagnosis labels mentioned in the dialogue were masked for the clinician (Appendix~\ref{app:data_prep}). These decisions form the \emph{Clinician} series in Figure~\ref{fig:human_eval}(a) and were scored using the same metrics as the models (Section~\ref{sec:evaluation_metrics}).

\paragraph{Stage 2: trajectory rating.}
After completing Stage~1, the clinician reviewed each \textsc{DoC} (SFT+RL, seed 42) trajectory together with the encounter audio (Figure~\ref{fig:study-ui-stage2}). With the reference label visible, the clinician assigned one overall rating---helpful, neutral, or harmful---and recorded one primary reason and detailed comments for each non-helpful rating. Never-committing trajectories were also rated by the clinician. Because reference labels were visible, these ratings reflect informed post-hoc judgments of perceived usefulness of \textsc{DoC}.

\paragraph{Case studies.}
Figures~\ref{fig:error-case-1}--\ref{fig:error-case-6} present six illustrative trajectories and their clinician assessments. Each figure gives the reference label, model prediction, and clinician rating, reason, and comment. Shaded rows show \textsc{DoC}'s decisions in the transcript. These cases illustrate concerns about diagnostic accuracy, commitment timing, non-commitment, and post-commit instability.

\subsection{Prompts}
\label{app:prompt}

We use diagnosis-only prompting for Phase~1 boundary estimation and \textsc{Defer}/\textsc{Commit} prompting for Phase~2 SFT and RL. The latter is also used for standard evaluation, the zero-shot in-context baselines (Qwen3-1.7B/4B, Doctor-R1, Claude Opus 5, and GPT-6 Sol), and the diagnosis components of ASPIRE, CaRT, and R-Tuning. The forced-commit pass uses assistant-output prefilling, as described below.

The label set lists the full dataset vocabulary in ID order: 35 protocols for EMSDialog and 33 diagnoses for MediTOD. Each dialogue prefix lists its utterances with their utterance numbers, speaker roles, and text. For all in-context baselines, the chat template is applied with \textbf{thinking disabled} for the real-time conversation setting. Dialogue prefixes follow the sampling procedure in Section~\ref{sec:setup}.

\paragraph{Phase~1: diagnosis-only prompting.}
During boundary estimation, the labeler is instructed to select exactly one diagnosis from $\mathcal{Y}$ for each dialogue prefix; \textsc{Defer} is not an available option. Both the initial and subsequent labelers are trained using diagnosis labels only. The correctness rate in Eq.~\eqref{eq:prefix_correctness} is therefore computed from diagnosis samples, without a deferral decision.

\paragraph{System prompt for \textsc{Defer}/\textsc{Commit} prediction.}
The following prompt carries the explicit-abstention-option design of MedAbstain~\cite{machcha-etal-2026-knowing} over to incremental dialogue by exposing \textsc{Defer} and \textsc{Commit} as the two actions over the full label set.

\begin{lstlisting}[style=promptstyle]
You are a clinical AI assistant supporting real-time diagnosis classification during an ongoing conversation.

Based only on the dialogue observed so far and the provided label set, choose exactly one action: DEFER or COMMIT.

1. Review the available symptoms, patient history, and other clinical evidence.
2. Choose DEFER if the evidence is insufficient, ambiguous, or conflicting, or if you cannot confidently distinguish a single diagnosis from the available labels.
3. Choose COMMIT only when the available evidence supports diagnosis with high confidence.
Output format:
4. For DEFER, set both "index" and "diagnosis" to null.
5. For COMMIT, set "index" to an integer label ID from the provided label set and "diagnosis" to its exact corresponding label text

{"action":"DEFER","index":null,"diagnosis":null}
or
{"action":"COMMIT","index":<label id>,"diagnosis":"<exact label text>"}
\end{lstlisting}

\paragraph{User prompt.}

\begin{lstlisting}[style=promptstyle]
Label set:
0: <label 0>
1: <label 1>
...

Dialogue so far:
[Turn 1] <role>: <utterance>
[Turn 2] <role>: <utterance>
...
[Turn <number>] <role>: <utterance>

Return the JSON object now:
\end{lstlisting}

\paragraph{Targets and outputs.}
For Phase~2 SFT, each prefix receives a \textsc{Defer} or diagnosis target according to Eq.~\eqref{eq:sft_target}. These targets and the model's generated actions use single-line JSON objects, shown below with line wrapping for display.

\begin{lstlisting}[style=promptstyle]
{"action":"DEFER","index":null,"diagnosis":null}
{"action":"COMMIT","index":<id>,"diagnosis":"<label text>"}
\end{lstlisting}

\paragraph{Forced-commit pass.}
The forced diagnosis used by Acc@Def is obtained using the \textsc{Defer}/\textsc{Commit} prompt above, with the assistant output prefilled as follows:

\begin{lstlisting}[style=promptstyle]
{"action":"COMMIT","index":
\end{lstlisting}

This fixes the action field to \textsc{Commit} while allowing the model to generate the diagnosis.

\subsection{Notation}
\label{app:notation}

Table~\ref{tab:notation} summarizes the notation of
Sections~\ref{sec:problem_formulation}--\ref{sec:setup}.

\begin{table*}[t]
\centering
\small
\setlength{\tabcolsep}{6pt}
\renewcommand{\arraystretch}{1.12}
\begin{tabular}{@{}l p{12.2cm}@{}}
\toprule
\textbf{Symbol} & \textbf{Definition} \\
\midrule
\multicolumn{2}{@{}l}{\textbf{Dialogues and actions}} \\
$i$, $N_{\mathrm{tr}}$, $N_{\mathrm{val}}$ &
Dialogue index within each split, and the numbers of training
and validation dialogues \\
$t$, $T_i$ & Prefix index, $t=1,\ldots,T_i$, and the number of prefixes presented to the model for dialogue $i$ \\
$x_{i,t}$ & The $t$-th dialogue prefix presented to the model; $x_{i,T_i}$ contains the complete dialogue \\
$\mathcal{D}_0$ & Labeled dataset of complete dialogues, $\{(x_{i,T_i},y_i)\}_{i=1}^{N_{\mathrm{tr}}}$ \\
$\mathcal{Y}$, $y_i$ & Diagnosis label set and ground-truth diagnosis of dialogue $i$ \\
\textsc{Defer} & Action that withholds a diagnosis pending further evidence \\
$\mathcal{A}$, $a_{i,t}$ & Action space $\mathcal{Y}\cup\{\textsc{Defer}\}$ and the action produced for prefix $x_{i,t}$ \\
$z_{i,t}$, $\mathcal{Z}_{\mathrm{early}}$ & Topic of the utterance ending $x_{i,t}$ and the set of early-zone topics \\
$t_i^{\mathrm{e}}$ & Early-evidence boundary: index of the last prefix ending in an early-zone utterance (Eq.~\eqref{eq:early_boundary}); $t_i^{\mathrm{e}}=0$ if no such prefix exists \\
$t_i^{\mathrm{c}}$ & Index of the first prefix yielding a diagnosis, $\min\{t:a_{i,t}\neq\textsc{Defer}\}$; equals $T_i+1$ if the policy never commits \\
$t_i^*$ & Estimated commitment boundary: prefixes with $t<t_i^*$ are assigned \textsc{Defer} targets \\
$t_i^{\ell}$ & Lateness reference: index of the last prefix ending before the first explicit human diagnostic statement; $T_i$ if none \\
\addlinespace[3pt]
\multicolumn{2}{@{}l}{\textbf{Iterative commitment boundary estimation}} \\
$k$, $f_k$ & Boundary-estimation iteration, starting from $0$, and the labeler (diagnosis model) at iteration $k$ \\
$M$, $m$ & Number of diagnoses sampled per prefix ($M=8$) and sample index \\
$\hat{y}_{i,t}^{(k,m)}$ & The $m$-th diagnosis sampled from $f_k$ given $x_{i,t}$ \\
$q_{i,t}^{(k)}$ & Fraction of the $M$ samples that equal $y_i$ (Eq.~\eqref{eq:prefix_correctness}) \\
$\kappa$ & Reliability threshold that $q_{i,t}^{(k)}$ must reach and maintain\\
$t_i^{*(k)}$ & Estimated commitment boundary at iteration $k$ (Eq.~\eqref{eq:pseudo_boundary}) \\
$\mathcal{D}_{k+1}$ & Diagnosis-only prefix dataset used to train $f_{k+1}$ (Eq.~\eqref{eq:diagnosis_prefix_data}) \\
$\mathrm{Acc}_{\mathrm{val}}(f_k)$, $\bar p_{\mathrm{val}}^{*(k)}$ &
Dialogue validation accuracy and mean normalized commitment
boundary on the validation split \\
$\phi$ & Boundary fraction of the fixed-fraction control, $t_i^*=\phi\,T_i$ (Appendix~\ref{app:fixed_boundary}) \\
$\delta$ & Required minimal boundary advance per iteration (Eq.~\eqref{eq:pseudo_convergence}) \\
$a_{i,t}^*$ & Supervised target action for prefix $x_{i,t}$ (Eq.~\eqref{eq:sft_target}) \\
$\mathcal{D}_{\mathrm{SFT}}$, $\pi_{\mathrm{SFT}}$ & SFT dataset of (prefix, target action) pairs and the resulting initialized policy \\
\addlinespace[3pt]
\multicolumn{2}{@{}l}{\textbf{Reinforcement learning refinement}} \\
$G$, $g$ & Rollouts per dialogue and rollout index; $g$ is omitted when a single rollout is discussed \\
$a_{i,t}^{(g)}$, $\tau_i^{(g)}$ & Action for prefix $x_{i,t}$ in rollout $g$ and the full action sequence $(a_{i,1}^{(g)},\ldots,a_{i,T_i}^{(g)})$ \\
$P$ & Penalty for committing at or before $t_i^{\mathrm{e}}$ \\
$\alpha_1$, $\alpha_2$ & Deferral cost accrued over $(t_i^{\mathrm{e}},t_i^*]$ and over $(t_i^*,T_i]$, respectively \\
$r_{+}$, $r_{-}$, $r_{\varnothing}$ & Accuracy reward for a correct, an incorrect, and no first commit \\
$R_{\mathrm{time},i}$, $R_{\mathrm{acc},i}$, $R_{\mathrm{stab},i}$ & Timeliness, first-commit accuracy, and post-commit stability rewards (Table~\ref{tab:reward_cases}) \\
$R_i^{(g)}$, $\bar{R}_i$, $\sigma_i$ & Total reward of rollout $g$ and its mean and standard deviation over the $G$ rollouts \\
$A_i^{(g)}$, $\eta$ & Group-relative advantage and a small constant for numerical stability \\
$\pi_\theta$, $\beta$ & Policy being optimized and the coefficient of its KL penalty toward $\pi_{\mathrm{SFT}}$ \\
$\pi_{\mathrm{RL}}$ & Policy obtained after reinforcement learning refinement \\
\multicolumn{2}{@{}l}{\textbf{Evaluation}} \\
$t_i^{\mathrm{fc}}$ & Index of the first prefix with a correct diagnosis, used for Earliness; $t_i^{\mathrm{fc}}\geq t_i^{\mathrm{c}}$, with equality when the first commitment is correct \\
$\rho\in(0,1]$ & Target fraction of a dialogue's total utterance count used to evaluate Acc@$\rho$ \\
$C_i$ & Number of prediction changes after the first commitment \\
$\tilde{a}_{i,t}$ & Forced diagnosis at prefix $x_{i,t}$ with \textsc{Defer} disabled \\
\bottomrule
\end{tabular}
\caption{Notation.}
\label{tab:notation}
\end{table*}

\clearpage

\begin{algorithm*}[t]
\caption{Iterative Commitment Boundary Estimation}
\label{alg:boundary_estimation}
\small
\begin{algorithmic}[1]
\Require Complete training dataset $\mathcal{D}_0$
with $N_{\mathrm{tr}}$ labeled dialogues;
$N_{\mathrm{val}}$ labeled validation dialogues;
early-evidence boundaries $\{t_i^{\mathrm{e}}\}$ for both splits;
sample count $M$ and thresholds $\kappa,\delta$.
\Ensure Final training-set commitment boundaries
$\{t_i^*\}_{i=1}^{N_{\mathrm{tr}}}$.

\Statex \textit{Initialize the labeler and boundaries.}
\State Train diagnosis-only labeler $f_0$ on $\mathcal{D}_0$.
\State $k \gets 0$
\State $\{t_i^{*(k)}\}_{i=1}^{N_{\mathrm{tr}}} \gets$
\Call{EstimateBoundaries}{$f_k$, \texttt{tr}}
\State $\{t_i^{*(k)}\}_{i=1}^{N_{\mathrm{val}}} \gets$
\Call{EstimateBoundaries}{$f_k$, \texttt{val}}
\State Compute $\mathrm{Acc}_{\mathrm{val}}(f_k)$
on the complete validation dialogues.
\State $\bar{p}_{\mathrm{val}}^{*(k)}
\gets
\frac{1}{N_{\mathrm{val}}}
\sum_{i=1}^{N_{\mathrm{val}}}
\frac{t_i^{*(k)}}{T_i}$
\Comment{Using validation boundaries}

\Statex \textit{Propose and validate updates.}
\While{\textbf{true}}
    \State $\mathcal{D}_{k+1}
    \gets
    \displaystyle\bigcup_{i=1}^{N_{\mathrm{tr}}}
    \left\{
    (x_{i,t},y_i):
    t_i^{*(k)}\leq t\leq T_i
    \right\}$
    \Comment{Using training boundaries}

    \State Train candidate diagnosis-only labeler
    $f_{k+1}$ on $\mathcal{D}_{k+1}$.

    \State $\{t_i^{*(k+1)}\}_{i=1}^{N_{\mathrm{tr}}} \gets$
    \Call{EstimateBoundaries}{$f_{k+1}$, \texttt{tr}}
    \State $\{t_i^{*(k+1)}\}_{i=1}^{N_{\mathrm{val}}} \gets$
    \Call{EstimateBoundaries}{$f_{k+1}$, \texttt{val}}

    \State Compute $\mathrm{Acc}_{\mathrm{val}}(f_{k+1})$
    on the complete validation dialogues.
    \State $\bar{p}_{\mathrm{val}}^{*(k+1)}
    \gets
    \frac{1}{N_{\mathrm{val}}}
    \sum_{i=1}^{N_{\mathrm{val}}}
    \frac{t_i^{*(k+1)}}{T_i}$
    \Comment{Using validation boundaries}

    \If{$\mathrm{Acc}_{\mathrm{val}}(f_{k+1})
        \geq \mathrm{Acc}_{\mathrm{val}}(f_k)$
        \textbf{and}
        $\bar{p}_{\mathrm{val}}^{*(k+1)}
        -\bar{p}_{\mathrm{val}}^{*(k)}<-\delta$}
        \State $k \gets k+1$
        \Comment{Accept (Eq.~\eqref{eq:pseudo_convergence})}
    \Else
        \State $t_i^* \gets t_i^{*(k)}$
        for $i=1,\ldots,N_{\mathrm{tr}}$
        \Comment{Keep the last accepted training boundaries}
        \State \Return
        $\{t_i^*\}_{i=1}^{N_{\mathrm{tr}}}$
    \EndIf
\EndWhile

\Statex
\Function{EstimateBoundaries}{$f_k$, \textit{split}}
    \Statex \textit{Eqs.~\eqref{eq:prefix_correctness}--\eqref{eq:pseudo_boundary}
    for the given labeler; dialogue indices are local to the specified split.}

    \For{each dialogue $i$ in the specified split}
        \For{$t=t_i^{\mathrm{e}}+1,\ldots,T_i$}
            \State Sample
            $\hat{y}_{i,t}^{(k,1)},\ldots,
            \hat{y}_{i,t}^{(k,M)}$
            from $f_k(\cdot\mid x_{i,t})$.
            \State $q_{i,t}^{(k)}
            \gets
            \frac{1}{M}
            \sum_{m=1}^{M}
            \mathds{1}\!\left[
            \hat{y}_{i,t}^{(k,m)}=y_i
            \right]$
        \EndFor
        \State $t_i^{*(k)}\gets
        \min\Bigl(
        \bigl\{t>t_i^{\mathrm{e}}:
        q_{i,t'}^{(k)}\geq\kappa\ \ \forall\, t'\in[t,T_i]\bigr\}
        \cup\{T_i\}
        \Bigr)$
        \Comment{$T_i$ if no prefix qualifies}
    \EndFor

    \State \Return $\{t_i^{*(k)}\}$ for the specified split
\EndFunction

\end{algorithmic}
\end{algorithm*}

\clearpage

\definecolor{modelc}{HTML}{1F4E79}\definecolor{clinc}{HTML}{B34700}
\definecolor{modelrow}{HTML}{E8EFF7}\definecolor{casebg}{HTML}{F5F5F3}\definecolor{casebox}{HTML}{BFBFBF}
\providecommand{\modeltag}[1]{\leavevmode{\color{modelc}\textbf{M}\ifx\relax#1\relax\else:\,#1\fi}}
\providecommand{\clintag}[1]{\leavevmode{\color{clinc}\textbf{C}\ifx\relax#1\relax\else:\,#1\fi}}
\providecommand{\modelonly}[1]{\leavevmode{\color{modelc}#1}}
\providecommand{\gaprow}[3]{\multicolumn{4}{@{}p{\textwidth}@{}}{\hspace{1.2em}{\itshape\color{black!55}[\ldots]\ utterances #1--#2 omitted#3}}\\}
\definecolor{dcorrc}{HTML}{1F4E79}\definecolor{dwrongc}{HTML}{B2182B}\definecolor{ddeferc}{HTML}{8C8C8C}
\providecommand{\dcorr}[1]{\leavevmode{\upshape\color{dcorrc}#1~\ding{51}}}
\providecommand{\dwrong}[1]{\leavevmode{\upshape\color{dwrongc}#1~\ding{55}}}
\providecommand{\ddefer}{\leavevmode{\upshape\color{ddeferc}\textsc{defer}}}

\begin{figure*}[p]
\centering\footnotesize
\fcolorbox{casebox}{casebg}{\parbox{\dimexpr\textwidth-2\fboxsep-2\fboxrule\relax}{\raggedright
\textbf{Gold:} heat exposure/heat stroke \quad \textbf{Dialogue:} D17, 31 utterances\\[2pt]\textbf{Model:} heat exposure/heat stroke~\ding{51} \\[2pt]\textbf{Clinician's rating}: harmful; \textbf{Reason}: too early. \textbf{Comment}: \emph{The model made the decision too early, because a patient saying that they are burning up and shivering could have something other than a heat stroke (like an infection). Also the model confirmed heat stroke without hearing if the patient was outside and for how long}\\[2pt]{\color{black!60}Model output for each evaluated prefix: {\color{dcorrc}correct~\ding{51}}, {\color{dwrongc}wrong~\ding{55}}, {\color{ddeferc}\textsc{defer}}; shaded rows mark transitions to a diagnosis label; transitions to \textsc{Defer} are unshaded. FR~=~first responder, Pt~=~patient, By~=~bystander; [MASK]~=~diagnosis masked in the study.}}}
\par\medskip
\begin{tabularx}{\textwidth}{@{}r@{\hspace{5pt}}l@{\hspace{6pt}}X@{\hspace{6pt}}p{3.9cm}@{}}
\toprule
No. & & Utterance & Model Output to Clinician \\
\midrule
1 & Pt & Man, I'm terribly sick. Can you help me? &  \\
2 & FR1 & What seems to be wrong? What what are you feeling like right now? & \ddefer{} \\
3 & Pt & I feel like I'm burning up and I'm shivering at the same time. &  \\
\rowcolor{modelrow}4 & FR1 & Well you you appear to be a little sunburn. How long you been outside? & \dcorr{heat stroke} \\
5 & Pt & Um since early this morning, I don't know, what time is it now? &  \\
6 & FR1 & You been drinking plenty of fluids today? & \dcorr{heat stroke} \\
7 & Pt & No. No. I haven't. Been walking most of the day. &  \\
8 & FR1 & We're gonna take some vital signs real quick, see what they're telling us. & \dcorr{heat stroke} \\
9 & Pt & Okay. &  \\
10 & FR1 & Your blood pressure is 113 over 79. & \dcorr{heat stroke} \\
11 & Pt & Is that good or bad? &  \\
12 & FR1 & That's pretty good. & \dcorr{heat stroke} \\
13 & Pt & Okay. I still feel awful. &  \\
14 & FR1 & Your respirations are 20 so that's good. Your pulse your pulse is a little high, it's 135, but it's regular. & \dcorr{heat stroke} \\
15 & Pt & All right. &  \\
16 & FR1 & Your oxygen's at 97, so that's good. You're gonna feel a little prick we're gonna check your sugar, too. Sorry to prick your finger real quick. & \dcorr{heat stroke} \\
17 & Pt & Oh. Oh. &  \\
18 & FR1 & I'm sorry. That's 109, so that's good, too. & \dcorr{heat stroke} \\
19 & Pt & Well if if all this is so good how come I how come I feel like this? &  \\
20 & FR1 & You've probably been outside a little too long today, so you need to get some fluids we need to get some fluids in you. You probably should take you to the hospital and get checked out. & \dcorr{heat stroke} \\
\gaprow{21}{31}{ --- model: \dcorr{heat stroke} after utterances 22, \ldots, 31}
\bottomrule
\end{tabularx}
\caption[Case 1: harmful, too early]{\textbf{Case 1 (harmful: too early)}
\label{fig:error-case-1}}
\end{figure*}

\begin{figure*}[p]
\centering\footnotesize
\fcolorbox{casebox}{casebg}{\parbox{\dimexpr\textwidth-2\fboxsep-2\fboxrule\relax}{\raggedright
\textbf{Gold:} stroke/TIA \quad \textbf{Dialogue:} D28, 237 utterances\\[2pt]\textbf{Model:} head injury~\ding{55} \\[2pt]\textbf{Clinician's rating}: harmful; \textbf{Reason}: wrong prediction. \textbf{Comment}: \emph{The model chose head injury based on the patient's description of the worst headache, but a severe headache alone does not mean that a head injury has occurred. There was not enough evidence of an injury to support that choice. The crew later noted left-sided deficits and performed stroke assessments, so this prediction could have pointed the assessment in the wrong direction}.\\[2pt]{\color{black!60}Model output for each evaluated prefix: {\color{dcorrc}correct~\ding{51}}, {\color{dwrongc}wrong~\ding{55}}, {\color{ddeferc}\textsc{defer}}; shaded rows mark transitions to a diagnosis label; transitions to \textsc{Defer} are unshaded. FR~=~first responder, Pt~=~patient, By~=~bystander; [MASK]~=~diagnosis masked in the study.}}}
\par\medskip
\begin{tabularx}{\textwidth}{@{}r@{\hspace{5pt}}l@{\hspace{6pt}}X@{\hspace{6pt}}p{3.9cm}@{}}
\toprule
No. & & Utterance & Model Output to Clinician \\
\midrule
1 & FR2 & Alright [REDACTED], it looks like we just got dispatched to a 32-year-old female who having the worst pain in her head that she's ever had. &  \\
2 & FR1 & Headache. & \ddefer{}\\
3 & FR1 & Ooh, could be a [MASK]. & \\
\gaprow{4}{17}{ --- model: \ddefer{}}
\rowcolor{modelrow}18 & Pt & My head hurts so bad. & \dwrong{head injury} \\
19 & Pt & This is the worst I've ever had. &  \\
20 & FR2 & Worst headache you've ever had? & \dwrong{head injury} \\
\gaprow{21}{47}{ --- model: \dwrong{head injury} after utterances 22, \ldots, 46}
48 & FR1 & Pulse of 118 and uh oxygen sats at 100. & \dwrong{head injury} \\
49 & FR2 & Okay, left sided deficit on the hand. &  \\
\gaprow{50}{51}{ --- model: \dwrong{head injury} after utterance 50}
52 & FR2 & Okay, can you say, can't teach an old dog new tricks? & \dwrong{head injury} \\
53 & Pt & Can't teach dog tricks. &  \\
\gaprow{54}{57}{ --- model: \dwrong{head injury} after utterances 54, 56}
58 & FR2 & Got some left sided deficit as well. & \dwrong{head injury} \\
\gaprow{59}{62}{ --- model: \dwrong{head injury} after utterances 60, 62}
63 & FR2 & We do want to get to the hospital as quickly as possible 'cause this could possibly be a [MASK]. &  \\
\gaprow{64}{71}{ --- model: \dwrong{head injury} after utterances 64, \ldots, 70}
\rowcolor{modelrow}72 & FR1 & Yes, sir. & \dcorr{stroke/TIA} \\
\gaprow{73}{149}{ --- model: \dcorr{stroke/TIA} after utterances 74, \ldots, 148}
150 & FR2 & 117 on the blood glucose. & \dcorr{stroke/TIA} \\
151 & FR1 & Okay, so sugar's normal. &  \\
152 & FR1 & We have done BEFAST, we have done RACE. & \dcorr{stroke/TIA} \\
153 & FR1 & We're going to take her to [ADDRESS] General. &  \\
\gaprow{154}{237}{ --- model: \dcorr{stroke/TIA} after utterances 154, \ldots, 237}
\bottomrule
\end{tabularx}
\caption[Case 2: harmful, wrong prediction]{\textbf{Case 2 (harmful: wrong prediction)}
\label{fig:error-case-2}}
\end{figure*}

\begin{figure*}[p]
\centering\footnotesize
\fcolorbox{casebox}{casebg}{\parbox{\dimexpr\textwidth-2\fboxsep-2\fboxrule\relax}{\raggedright
\textbf{Gold:} diabetic -- hypoglycemia \quad \textbf{Dialogue:} D30, 54 utterances\\[2pt]
\textbf{Model:} \textsc{Defer}~\ding{55}\\[2pt]
\textbf{Clinician's rating:} harmful; \textbf{Reason:} never commits; \textbf{Comment:} \emph{the evidence for hypoglycemia accumulates steadily: the bystander reports that the crew elicits sugar problems, notes warm and moist skin with unremarkable vital signs, measures a blood sugar of 52 and gives oral glucose. However the model defers at every step and gives no recommendation.}\\[2pt]
{\color{black!60}Model output for each evaluated prefix: {\color{dcorrc}correct~\ding{51}}, {\color{dwrongc}wrong~\ding{55}}, {\color{ddeferc}\textsc{defer}}; shaded rows mark transitions to a diagnosis label; transitions to \textsc{Defer} are unshaded. FR~=~first responder, Pt~=~patient, By~=~bystander.}}}
\par\medskip
\begin{tabularx}{\textwidth}{@{}r@{\hspace{5pt}}l@{\hspace{6pt}}X@{\hspace{6pt}}p{3.9cm}@{}}
\toprule
No. & & Utterance & Model Output to Clinician \\
\midrule
\gaprow{1}{5}{ --- model: \ddefer{} after utterances 2, 4}
6 & FR1 & Hello, Fire Rescue! & \ddefer{} \\
\gaprow{7}{8}{ --- model: \ddefer{} after utterance 8}
9 & FR1 & I understand you're not feeling well today. &  \\
10 & FR1 & Okay. & \ddefer{} \\
11 & FR1 & You're not feeling well. What's going on with you? &  \\
\gaprow{12}{13}{ --- model: \ddefer{} after utterance 12}
14 & FR1 & Is he normally like this? & \ddefer{} \\
15 & By1 & No, he he's acting a bit strangely today. I couldn't get a hold of him. So, I decided to come over here. &  \\
16 & FR1 & Okay. Do you know any What's your past medical history? You got uh heart disease or respiratory problems or uh diabetes or anything like that? You Do you take any medicines? & \ddefer{} \\
17 & FR1 & Sugar problems. Okay. Does he take any medicines? &  \\
\gaprow{18}{26}{ --- model: \ddefer{} after utterances 18, \ldots, 26}
27 & FR1 & Yeah, he is a little warm, a little moist. &  \\
\gaprow{28}{32}{ --- model: \ddefer{} after utterances 28, 30, 32}
33 & FR1 & Okay. Pulse ox is 98. &  \\
34 & FR1 & Your Your pulse is 67. Okay? & \ddefer{} \\
\gaprow{35}{37}{ --- model: \ddefer{} after utterance 36}
38 & FR1 & Okay. All right. Let's see. Okay. Your BP is 110 over 70. You know what your blood pressure normally is? & \ddefer{} \\
39 & FR1 & Okay. All right. So, you You said you got a sugar problem? &  \\
40 & FR1 & Do you think so? Okay. All right. & \ddefer{} \\
41 & FR1 & I'm going to go ahead, and I'm going to go ahead and check your sugar level, okay? &  \\
\gaprow{42}{47}{ --- model: \ddefer{} after utterances 42, 44, 46}
48 & FR1 & Okay. Your blood sugar level's 52. Sir, can Can Let me see if you can swallow. Can you swallow, sir? & \ddefer{} \\
49 & FR1 & Okay. Good job. Good job. I'm going to give you a little something over here. &  \\
\gaprow{50}{52}{ --- model: \ddefer{} after utterances 50, 52}
53 & FR1 & All right. Now, don't swallow it. Just kind of put it right in your right in your cheek, okay? Right in your cheek and All right. All right. That's a good job. That's really good. Okay. Here. Let me get put some more. All right. Okay. All right. Don't swallow it now. Don't swallow. I just want you [\ldots] &  \\
54 & FR1 & You okay? Okay. That's fine. You just sit right there. Now, we're going to get the ambulance in here in just a minute. Okay? Now, you do need to go to the hospital, okay? Let me follow back up and check your check your BP again. Okay? [\ldots] & \ddefer{} \\
\bottomrule
\end{tabularx}
\caption[Case 3: harmful, never commits]{\textbf{Case 3 (harmful: never commits)}\label{fig:error-case-3}}
\end{figure*}

\begin{figure*}[p]
\centering\footnotesize
\fcolorbox{casebox}{casebg}{\parbox{\dimexpr\textwidth-2\fboxsep-2\fboxrule\relax}{\raggedright
\textbf{Gold:} chest pain -- cardiac suspected \quad \textbf{Dialogue:} D42, 333 utterances\\[2pt]\textbf{Model:} chest pain -- cardiac suspected~\ding{51} \\[2pt]\textbf{Clinician's rating}: harmful; \textbf{Reason}: first correct, switches to wrong. \textbf{Comment}: \emph{The model initially selected the correct protocol but repeatedly switched to cardiac arrest. These changes made its guidance difficult to rely on and reduced the usefulness of the initially correct prediction}\\[2pt]{\color{black!60}Model output for each evaluated prefix: {\color{dcorrc}correct~\ding{51}}, {\color{dwrongc}wrong~\ding{55}}, {\color{ddeferc}\textsc{defer}}; shaded rows mark transitions to a diagnosis label; transitions to \textsc{Defer} are unshaded. FR~=~first responder, Pt~=~patient, By~=~bystander.}}}
\par\medskip
\begin{tabularx}{\textwidth}{@{}r@{\hspace{5pt}}l@{\hspace{6pt}}X@{\hspace{6pt}}p{3.9cm}@{}}
\toprule
No. & & Utterance & Model Output to Clinician \\
\midrule
1 & FR1 & Are we set to go? &  \\
2 & FR2 & Yeah, we're set to go. & \ddefer{} \\
3 & FR1 & All right, it looks like we have a patient with chest pain. &  \\
\gaprow{4}{37}{ --- model: \ddefer{}}
38 & FR2 & Yep. & \ddefer{} \\
39 & FR1 & We're going to do a couple of things to check you out and take care of you here. &  \\
\rowcolor{modelrow}40 & FR1 & All right. & \dcorr{chest pain (cardiac)} \\
\gaprow{41}{77}{ --- model: \dcorr{chest pain (cardiac)} after utterances 42, \ldots, 76}
78 & Pt & An elephant sitting on my chest. & \dcorr{chest pain (cardiac)} \\
79 & FR1 & Yeah, is it like a heavy pressure? &  \\
80 & Pt & Yeah, it's just it kind of feels like the whole day's gotten me down, you know. & \ddefer{} \\
81 & FR1 & Can we get a 12-lead on him? &  \\
82 & FR1 & The 12-lead is in the back with the leads. & \ddefer{} \\
83 & FR1 & And does it radiate anywhere or is it just in the middle of the chest? &  \\
84 & Pt & Yeah, I gotta be honest, it got worse when I started smoking. & \ddefer{} \\
85 & Pt & I thought it would have helped but it's kind of &  \\
86 & FR1 & No, it made it worse. & \ddefer{} \\
87 & Pt & Yeah, it's kind of all over my chest and &  \\
88 & FR2 & Alright my friend, we're going to get a picture of your heart real quick, okay? & \ddefer{} \\
89 & FR1 & All right. &  \\
\rowcolor{modelrow}90 & FR1 & Yeah. Scale of 1 to 10, 10's like your arm being cut off. & \dcorr{chest pain (cardiac)} \\
\gaprow{91}{291}{ --- model: \ddefer{} after utterance 92; \dcorr{chest pain (cardiac)} after utterances 94, \ldots, 116; \ddefer{} after utterance 118; \dcorr{chest pain (cardiac)} after utterance 120; \ddefer{} after utterance 122; \dcorr{chest pain (cardiac)} after utterances 124, 126, 128; \ddefer{} after utterance 130; \dcorr{chest pain (cardiac)} after utterance 132; \ddefer{} after utterance 134; \dcorr{chest pain (cardiac)} after utterances 136, \ldots, 152; \ddefer{} after utterances 154, \ldots, 208; \dcorr{chest pain (cardiac)} after utterances 210, \ldots, 290}
\rowcolor{modelrow}292 & Pt & The helicopter? The airplane? & \dwrong{cardiac arrest} \\
\gaprow{293}{298}{ --- model: \dwrong{cardiac arrest} after utterances 294, 296, 298}
299 & FR1 & I'm going to give you some aspirin. &  \\
\rowcolor{modelrow}300 & Pt & Hey, don't and don't use my left arm. & \dcorr{chest pain (cardiac)} \\
\gaprow{301}{307}{ --- model: \dcorr{chest pain (cardiac)} after utterances 302, 304, 306}
\rowcolor{modelrow}308 & FR1 & Then I'm going to give you some uh morphine to help with that pain. & \dwrong{cardiac arrest} \\
\gaprow{309}{319}{ --- model: \dwrong{cardiac arrest} after utterances 310, \ldots, 318}
\rowcolor{modelrow}320 & Pt & Okay. & \dcorr{chest pain (cardiac)} \\
\gaprow{321}{333}{ --- model: \dcorr{chest pain (cardiac)} after utterances 322, \ldots, 333}
\bottomrule
\end{tabularx}
\caption[Case 4: harmful, first correct, switches to wrong]{\textbf{Case 4 (harmful: first correct, switches to wrong)}
\label{fig:error-case-4}}
\end{figure*}

\begin{figure*}[p]
\centering\footnotesize
\fcolorbox{casebox}{casebg}{\parbox{\dimexpr\textwidth-2\fboxsep-2\fboxrule\relax}{\raggedright
\textbf{Gold:} stroke/TIA \quad \textbf{Dialogue:} D05, 254 utterances\\[2pt]\textbf{Model:} stroke/TIA~\ding{51} \\[2pt]\textbf{Clinician's rating}: neutral; \textbf{Reason}: too late. \textbf{Comment}: \emph{It really recognized the stroke when the provider said that the patient might be having a [MASK]. This would be too late to be helpful as the provider is already concerned that the patient may be having a stroke}\\[2pt]{\color{black!60}Model output for each evaluated prefix: {\color{dcorrc}correct~\ding{51}}, {\color{dwrongc}wrong~\ding{55}}, {\color{ddeferc}\textsc{defer}}; shaded rows mark transitions to a diagnosis label; transitions to \textsc{Defer} are unshaded. FR~=~first responder, Pt~=~patient, By~=~bystander; [MASK]~=~diagnosis masked in the study.}}}
\par\medskip
\begin{tabularx}{\textwidth}{@{}r@{\hspace{5pt}}l@{\hspace{6pt}}X@{\hspace{6pt}}p{3.9cm}@{}}
\toprule
No. & & Utterance & Model Output to Clinician \\
\midrule
1 & FR1 & All right. &  \\
2 & FR2 & Oh. & \ddefer{} \\
3 & FR1 & Let's go to [REDACTED] now. &  \\
\gaprow{4}{15}{ --- model: \ddefer{} after utterances 4, \ldots, 14}
16 & FR1 & Hi, we're the rescue squad. & \ddefer{} \\
17 & FR1 & Hi everyone. &  \\
18 & FR1 & What's going on? & \ddefer{} \\
19 & FR1 & My name's [REDACTED]. &  \\
20 & By1 & So, we we were hanging out, we we were just, you know, eating pizza, you know, watching a movie and uh she she said she didn't feel well but it it came out a little slurred. & \ddefer{} \\
21 & By1 & Um, like she was having a hard time talking and now she can't really respond to me at all and I don't know what's going on. &  \\
\gaprow{22}{53}{ --- model: \ddefer{} after utterances 22, \ldots, 52}
54 & FR2 & All right, we do have left-sided deficit. & \ddefer{} \\
\gaprow{55}{57}{ --- model: \ddefer{} after utterance 56}
58 & FR2 & Can you do me a favor and smile real big? & \ddefer{} \\
59 & FR2 & Got left-sided deficit on the face as well. &  \\
\gaprow{60}{63}{ --- model: \ddefer{} after utterances 60, 62}
64 & FR2 & Okay, we got speech deficit as well. & \ddefer{} \\
\gaprow{65}{119}{ --- model: \ddefer{} after utterances 66, \ldots, 118}
120 & FR1 & I don't think it will be, but we're going to see if your blood sugar's low. & \ddefer{} \\
\gaprow{121}{123}{ --- model: \ddefer{} after utterance 122}
124 & FR1 & One, two, three, pop. & \ddefer{} \\
125 & FR1 & Okay. &  \\
126 & FR1 & And Accu-Chek comes back at 156. & \ddefer{} \\
\gaprow{127}{166}{ --- model: \ddefer{} after utterances 128, \ldots, 166}
167 & FR1 & Your rhythm's looking really good, for whatever that's worth. &  \\
168 & FR2 & All right, so it looks like you have normal sinus rhythm. & \ddefer{} \\
169 & FR2 & And it looks like the blood glucose is good. &  \\
\rowcolor{modelrow}170 & FR2 & So to me, that looks like you might be having a [MASK]. & \dcorr{stroke/TIA} \\
\gaprow{171}{254}{ --- model: \dcorr{stroke/TIA} after utterances 172, \ldots, 254}
\bottomrule
\end{tabularx}
\caption[Case 5: neutral, too late]{\textbf{Case 5 (neutral: too late)}\label{fig:error-case-5}}
\end{figure*}

\begin{figure*}[p]
\centering\footnotesize
\fcolorbox{casebox}{casebg}{\parbox{\dimexpr\textwidth-2\fboxsep-2\fboxrule\relax}{\raggedright
\textbf{Gold:} overdose/poisoning -- opioid \quad \textbf{Dialogue:} D41, 55 utterances\\[2pt]\textbf{Model:} overdose/poisoning -- opioid~\ding{51} \\[2pt]\textbf{Clinician's rating}: neutral; \textbf{Reason}: first correct, returns to defer. \textbf{Comment}: \emph{The model committed to an opioid overdose at beginning, after the bystander mentioned drug paraphernalia and the responder noted snoring respirations at about four breaths per minute. It returned to \textsc{defer} later, exactly while the evidence was getting stronger (pinpoint pupils, an oxygen saturation of 86), and re-committed to opioid overdose just before the crew named the opioid overdose. The diagnosis was never wrong, but the withdrawal made the guidance look uncertain for no reason.}\\[2pt]{\color{black!60}Model output for each evaluated prefix: {\color{dcorrc}correct~\ding{51}}, {\color{dwrongc}wrong~\ding{55}}, {\color{ddeferc}\textsc{defer}}; shaded rows mark transitions to a diagnosis label; transitions to \textsc{Defer} are unshaded. FR~=~first responder, Pt~=~patient, By~=~bystander.}}}
\par\medskip
\begin{tabularx}{\textwidth}{@{}r@{\hspace{5pt}}l@{\hspace{6pt}}X@{\hspace{6pt}}p{3.9cm}@{}}
\toprule
No. & & Utterance & Model Output to Clinician \\
\midrule
1 & By1 & Hey guys, I'm glad you're here. I don't know what happened. This, this guy, I found them on the, on the bathroom floor here when I, I don't know what's going on with them, but you can see all this drug stuff around though. &  \\
\gaprow{2}{5}{ --- model: \ddefer{} after utterances 2, 4}
\rowcolor{modelrow}6 & FR1 & Okay. So, it sounds like she's got some snoring respirations, like she's not really, not really breathing too hot, maybe only about four times a minute. [\ldots] & \dcorr{overdose (opioid)} \\
\gaprow{7}{9}{ --- model: \dcorr{overdose (opioid)} after utterance 8}
10 & FR1 & Alright. Alright, NPA is in. [\ldots] would you mind going ahead and getting uh, pulse ox reading on her, um, just sort of full set of vital signs? & \ddefer{} \\
11 & FR2 & Yes. &  \\
12 & FR1 & Alright, thank you very much. Gonna go ahead and look at her eyes. Patient's pinpoint pupils, probably only around uh, 2 millimeters or so. [\ldots] & \ddefer{} \\
13 & FR2 & It is 86. &  \\
\gaprow{14}{19}{ --- model: \ddefer{} after utterances 14, 16, 18}
\rowcolor{modelrow}20 & FR1 & Alright, with, blood pressure 140 over 70, that looks pretty good. & \dcorr{overdose (opioid)} \\
21 & FR2 & And her pulse is 110. &  \\
22 & FR1 & Pulse is 110, that's good, good and strong. Alright, so if we've got oxygen on there and all that, I think that uh, looks like we're just dealing with an [MASK] here. [\ldots] & \dcorr{overdose (opioid)} \\
\gaprow{23}{55}{ --- model: \dcorr{overdose (opioid)} after utterances 24, \ldots, 55}
\bottomrule
\end{tabularx}
\caption[Case 6: neutral, first correct, returns to defer]{\textbf{Case 6 (neutral: first correct, returns to defer)}\label{fig:error-case-6}}
\end{figure*}

\begin{figure*}[p]
\centering
{\setlength{\fboxsep}{0pt}\fcolorbox{casebox}{white}{\includegraphics[width=0.9\textwidth]{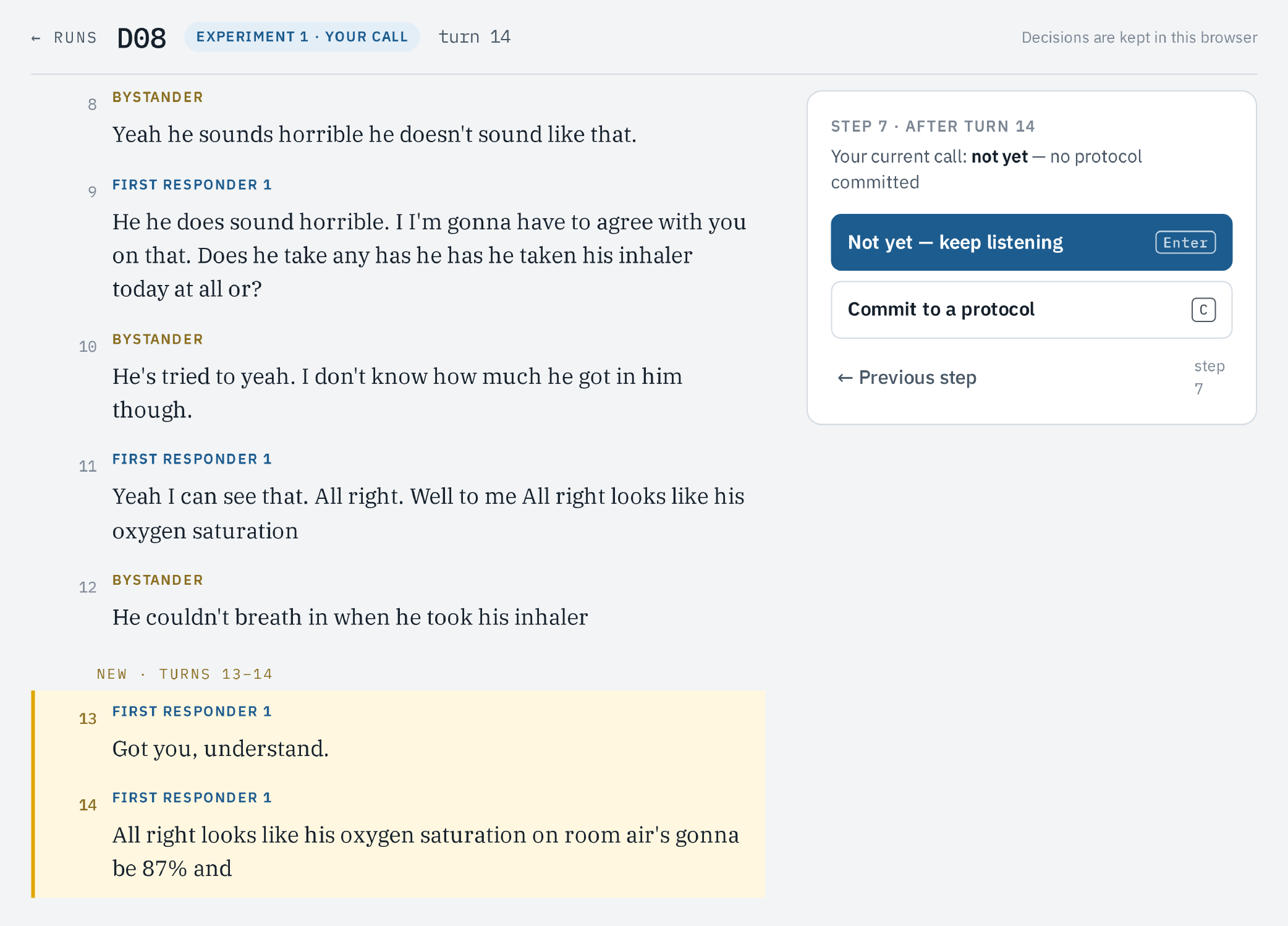}}}
\caption[Clinician study interface, Stage 1]{Stage~1 incremental-diagnosis interface for D08. At each prefix, the clinician selects \textsc{Defer} or one of 35 diagnosis labels; model outputs and the reference label are hidden, and diagnosis names are masked.}
\label{fig:study-ui-stage1}
\end{figure*}

\begin{figure*}[p]
\centering
{\setlength{\fboxsep}{0pt}\fcolorbox{casebox}{white}{\includegraphics[width=0.9\textwidth]{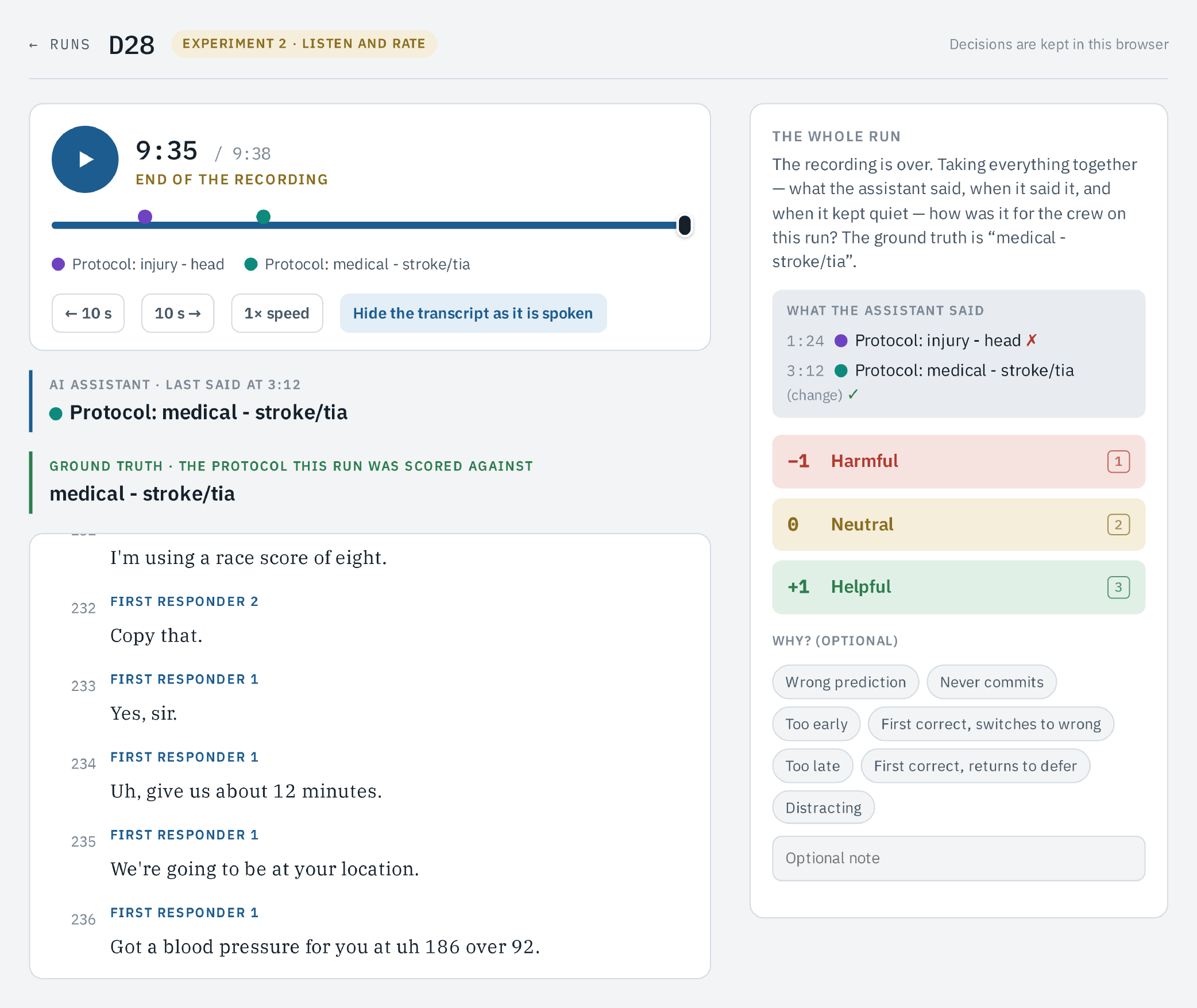}}}
\caption[Clinician study interface, Stage 2]{Stage~2 trajectory-rating interface for D28. The clinician reviews \textsc{DoC}'s trajectory with audio and a visible reference label, rates it helpful/neutral/harmful, and adds one primary reason and comments for non-helpful ratings.}
\label{fig:study-ui-stage2}
\end{figure*}

\end{document}